\documentclass{article}
\usepackage{arxiv}
\usepackage{authblk}
\usepackage{amsmath}
\usepackage{amssymb}
\usepackage{amsfonts}
\usepackage{amsthm}          
\newtheorem*{remark}{Remark} 
\theoremstyle{definition}
\newtheorem{definition}{Definition}
\usepackage{accents}
\usepackage{nicefrac}
\usepackage{tikz}
\usetikzlibrary{calc, arrows.meta, positioning, shapes.geometric, arrows}
\usepackage{xspace}
\usepackage{bm}
\usepackage{listofitems}
\usepackage{xcolor}
\usepackage{contour}
\usepackage{enumitem} 
\usepackage{booktabs}
\usepackage{soul}
\usepackage{arydshln}
\usepackage{graphicx,scalerel,centernot}
\usepackage{caption}
\usepackage{algorithm}
\usepackage{algpseudocode}
\usepackage{subfig}
\usepackage{float}
\usepackage[toc,page]{appendix}
\usepackage{threeparttable}
\usepackage[dvipsnames]{xcolor}
\usepackage{url}
\usepackage{adjustbox}

\usetikzlibrary{calc, arrows.meta, positioning, shapes.geometric, arrows}

\setlist[enumerate]{itemsep=0mm}

\usetikzlibrary{positioning, arrows.meta, calc, quotes} 
\contourlength{1.4pt}

\tikzset{>=latex} 
\colorlet{myred}{red!80!black}
\colorlet{myblue}{blue!80!black}
\colorlet{mygreen}{green!60!black}
\colorlet{myorange}{orange!70!red!60!black}
\colorlet{mydarkred}{red!30!black}
\colorlet{mydarkblue}{blue!40!black}
\colorlet{mydarkgreen}{green!30!black}
\tikzstyle{node}=[thick,circle,draw=myblue,minimum size=22,inner sep=0.5,outer sep=0.6]
\tikzstyle{node in}=[node,green!20!black,draw=mygreen!30!black,fill=mygreen!25]
\tikzstyle{node noise}=[node,white!20!black,draw=mygreen!30!black,fill=white!25]
\tikzstyle{node hidden}=[node,blue!20!black,draw=myblue!30!black,fill=myblue!20]
\tikzstyle{node convol}=[node,orange!20!black,draw=myorange!30!black,fill=myorange!20]
\tikzstyle{node out}=[node,red!20!black,draw=myred!30!black,fill=myred!20]
\tikzstyle{connect}=[thick,mydarkblue] 
\tikzstyle{connect arrow}=[-{Latex[length=4,width=3.5]},thin,mydarkblue,shorten <=0.5,shorten >=1]
\tikzset{ 
	node 1/.style={node in},
	node 2/.style={node hidden},
	node 3/.style={node out},
}

\newcommand{\f}{\mkern-2mu f\mkern-3mu}              
\newcommand{\ci}{\mathbin{\perp\!\!\!\perp}}
\newcommand{\nci}{\mathbin{\centernot{\perp\!\!\!\perp}}}
\newcommand\scalemath[2]{\scalebox{#1}{\mbox{\ensuremath{\displaystyle #2}}}} 

\newsavebox{\measurebox}

\newcommand{\methodname}{R\textsc{e}X\xspace}

\setlist[enumerate]{itemsep=0mm}
\setlist[itemize]{itemsep=0pt, parsep=0pt}

\title{\methodname: Causal discovery based on machine learning and explainability techniques}

\author[1,3]{Jes\'us Renero}
\author[1,3]{Roberto Maestre}
\author[2,3]{~Idoia~Ochoa}
\affil[1]{\footnotesize BBVA Madrid, Spain}
\affil[2]{\footnotesize Tecnun, University of Navarra, San Sebasti\'{a}n, Spain}
\affil[3]{\footnotesize Instituto de Ciencia de Datos e Inteligencia Artificil (DATAI), University of Navarra, Pamplona, Spain}

\begin{document}
\maketitle	

\begin{abstract}
    Explainable Artificial Intelligence (XAI) techniques hold significant potential for enhancing the causal discovery process, which is crucial for understanding complex systems in areas like healthcare, economics, and artificial intelligence. However, no causal discovery methods currently incorporate explainability into their models to derive the causal graphs.
    Thus, in this paper we explore this innovative approach, as it offers substantial potential and represents a promising new direction worth investigating. 
    Specifically, we introduce \methodname, a causal discovery method that leverages machine learning (ML) models coupled with explainability techniques, specifically Shapley values, to identify and interpret significant causal relationships among variables.
    
    Comparative evaluations on synthetic datasets comprising continuous tabular data reveal that \methodname outperforms state-of-the-art causal discovery methods across diverse data generation processes, including non-linear and additive noise models. Moreover, \methodname was tested on the Sachs single-cell protein-signaling dataset, achieving a precision of 0.952 and recovering key causal relationships with no incorrect edges. Taking together, these results showcase \methodname's effectiveness in accurately recovering true causal structures while minimizing false positive predictions, its robustness across diverse datasets, and its applicability to real-world problems. By combining ML and explainability techniques with causal discovery, \methodname bridges the gap between predictive modeling and causal inference, offering an effective tool for understanding complex causal structures. \methodname is publicly available at \url{https://github.com/renero/causalgraph}.

\end{abstract}



\section{Introduction} \label{sec:intro}

Causal discovery --the process of identifying cause-and-effect relationships from observational data-- is a pivotal challenge in artificial intelligence (AI) and machine learning (ML). Unveiling causal structures enables robust predictions, facilitates counterfactual reasoning, and enhances decision-making processes in complex systems \cite{pearl2009causality}. Traditional methods for causal discovery often rely on statistical tests for independence and structural equation modeling, which may not scale efficiently with high-dimensional data or effectively capture intricate non-linear relationships \cite{spirtes2000causation, PetersJanzingSchoelkopf17}.

In recent years, ML models, particularly deep learning architectures, have achieved remarkable success in predictive tasks. However, these models are typically considered ``black boxes'' due to their lack of interpretability. This opacity has led to a growing interest in explainable AI (XAI) techniques, with Shapley values emerging as a prominent method for interpreting model predictions \cite{lundberg2017unified}. Shapley values, grounded in cooperative game theory, provide a principled approach to attributing the contribution of each feature to the output of a model by quantifying the average marginal contribution of a feature across all possible subsets of features \cite{shapley1953value}.

Explainable Artificial Intelligence (XAI) refers to a broad set of techniques aimed at improving the transparency of ML models, commonly divided into two categories: (i) models that are interpretable by design, and (ii) post-hoc explanation methods that analyze complex, often opaque, predictive models after training. This work focuses specifically on post-hoc feature attribution. We rely on Shapley values~\cite{shapley1953value} for this purpose, due to their firm theoretical foundation —particularly their satisfaction of axioms such as efficiency, symmetry, and additivity~\cite{shapley1953value,Winter2002Shapley}— as well as the availability of efficient, model-specific implementations~\cite{lundberg2017unified,chen_true_2020}. It is important to clarify that feature attributions derived from XAI methods, including Shapley values, should not be interpreted as causal explanations~\cite{pearl2018book,chen2022explaining}.

While Shapley values offer valuable insights into feature importance within a model's predictive framework, the link between feature importance and causal influence is non-trivial. A high Shapley value for a feature indicates a strong impact on the model's predictions but does not necessarily imply a direct causal relationship with the target variable \cite{datta2016algorithmic,sundararajan2019many}. Moreover, correlations captured by the model may be confounded by hidden variables or represent spurious associations \cite{pearl2009causality}. Therefore, leveraging Shapley values for causal discovery requires careful consideration to avoid misleading inferences. 

Despite the advancements in both causal discovery and explainable AI, there is a notable absence of methods that integrate explainability techniques--specifically Shapley values--into the causal discovery process to extract causal graphs. This highlights a significant gap in the intersection of these two fields, offering a promising new direction.

In this paper, we propose a novel method coined \methodname that integrates Shapley values into the causal discovery process, aiming to uncover causal structures by interpreting feature attributions from ML models. By cautiously exploiting the conceptual connection between Shapley values and causal relationships, \methodname helps narrow down the set of features that play significant roles in the causal graph. This targeted approach allows us to focus on the most influential variables for further causal analysis, improving the efficiency and accuracy of causal discovery in complex datasets. \methodname (see Sections~\ref{ss:mathematical-foundation} and~\ref{ss:choice-of-shapley}) makes use of Shapley values under the assumptions of Causal Markov Condition (Definition~\ref{def:markov-property}) and faithfulness (Definition~\ref{def:faithfullness}) as a quantitative approximation of conditional dependence. This allows \methodname to identify candidate causal parents from observational data, not by treating the explanations as causal \textit{per se}, but by integrating XAI tools into a broader causal discovery framework.

We demonstrate \methodname's competitive performance on synthetic datasets (Appendix{\ref{app:synthetic-datasets}}) traditionally employed to evaluate new models, encompassing diverse generation processes, as well as on two real-world complex system datasets. This performance underscores its practical applicability and the advantages over existing methods in addressing real-world causal discovery challenges. 

By integrating explainability techniques into the causal discovery workflow \methodname aims to generate more robust causal graphs, which are fundamental for advancing from associative predictive modeling towards causal inference (i.e., elucidating cause-effect mechanisms and predicting interventional outcomes). The benefits of this specific integration within \methodname are twofold: first, it enhances the transparency of the causal discovery process itself, as the selection of potential causal links is guided by interpretable Shapley values that quantify feature contributions. Second, by focusing the search on these significant and interpretable contributions, \methodname can more accurately identify genuine causal relationships from observational data, leading to more reliable causal models.

\subsection{Related work} \label{ss:classical-approaches-to-causal-discovery}

Classical approaches to causal discovery are primarily divided into three families: constraint-based methods, score-based methods, and structural causal models (SCMs).

Constraint-based methods, such as the PC algorithm \cite{spirtes2000causation} and Fast Causal Inference (FCI) \cite{spirtes1999}, traditionally build causal structures by iteratively performing statistical conditional independence (CI) tests on the observational data to remove edges from an initially complete graph. These methods explore statistical dependencies in the data to determine causal relationships between variables. Their reliance on direct CI tests applied to the data for edge pruning differs from \methodname's approach. \methodname first employs machine learning models to predict each variable (as detailed in Section \ref{s:Models-training}) and then uses Shapley values (Section \ref{ss:mathematical-foundation}) derived from these models as a more holistic proxy for conditional dependence, assessing feature importance across numerous predictive contexts rather than through specific CI tests for initial graph construction.

Score-based methods, like Greedy Equivalence Search (GES) \cite{chickering2002optimal}, search for the model that best fits the data according to a scoring criterion. This approach involves evaluating different graph structures and selecting the one that optimizes an objective function, balancing model fit and complexity. This contrasts with \methodname’s methodology of constructing the graph from locally significant parent-child relationships identified via Shapley-based feature contributions (Algorithm \ref{alg:parent-selection}), rather than optimizing a global graph score.

SCMs provide a framework for representing and estimating causal relationships through equations that describe how variables influence one another \cite{pearl2009causality}. Within this category, the LiNGAM algorithm \cite{shimizu2006linear} exploits non-Gaussianity in data to identify causal structures in linear models, assuming linear, non-Gaussian, acyclic relationships. \methodname, however, uses flexible ML models (Section \ref{s:Models-training}) such as DFNs and GBTs that inherently capture non-linearities and does not depend on non-Gaussianity for its core Shapley-based parent selection, reserving additive noise model assumptions primarily for a later edge orientation stage (Section \ref{ss:directing-edges}).
Causal Additive Models (CAM) \cite{buhlmann2014cam} extend these methods by modeling nonlinear relationships using additive noise models, often utilizing techniques like penalized regression for graph estimation. While \methodname also accommodates non-linearities through its underlying ML regressors and employs an additive noise assumption for orientation, its parent identification process (Algorithm \ref{alg:parent-selection}) is driven by Shapley-based feature importance derived from these general regressors, not by fitting a specific global additive model structure with penalization for graph selection. However, these classical SCM approaches often require strong assumptions about the data-generating process and may struggle with complex nonlinearities inherent in real-world data \cite{PetersJanzingSchoelkopf17}.

Hybrid methods, such as Max-Min Hill-Climbing (MMHC) \cite{tsamardinos2006max}, combine elements of both cons\-tra\-int-based and score-based approaches, for instance, by using CI tests to restrict the search space for a subsequent score-based optimization phase. These methods aim to leverage the strengths of each approach. \methodname's integration of ML and explainability is distinct, focusing on using XAI-derived feature importances from predictive models as the primary input for parent discovery before specific graph optimization or CI testing steps.

Recent advancements have led to the application of machine learning techniques in causal discovery. Algorithms like NOTEARS \cite{zheng2018dags} formulate causal discovery as a continuous optimization problem by defining an algebraic acyclicity constraint, enabling gradient-based learning of the entire DAG structure simultaneously. This unified optimization differs from \methodname’s sequential workflow, which involves distinct stages for predictive modeling, Shapley-based parent discovery (Algorithm \ref{alg:dag-construction}), edge orientation (Section \ref{ss:directing-edges}), and cycle resolution (Section \ref{ss:final-dag}). Structural Agnostic Modeling (SAM) \cite{kalainathan2019sam}, in particular, employs adversarial learning to model causal structures without making strong assumptions about the data distribution. These methods improve scalability and can capture nonlinear relationships but may still face challenges related to the interpretability of the discovery process itself \cite{yu2019dag}, a domain \methodname specifically addresses through its inherent use of XAI techniques to guide graph construction.


\section{Preliminaries} \label{s:preliminaries}

Let \( \mathbf{X} = \{ X_1, X_2, \ldots, X_p \} \) represent a vector of $p$ continuous variables, with unknown joint probability distribution $P(\mathbf{X})$. The observational causal discovery setting considers $m$ independent and identically distributed (i.i.d.) samples drawn from $P(\mathbf{X})$. The goal is then to infer the \textit{causal} directed acyclic graph (DAG) \( \mathcal{G} = (V, E) \), where \( V \) is the set of variables and \( E \) is the set of directed edges \( (X_j \rightarrow X_i) \) representing the causal relationship ``X\(_j\) causes X\(_i\)''.

\paragraph*{Causal relation \emph{vs.} statistical dependence}
Throughout the paper we distinguish two notions:

\begin{enumerate}[label=(\roman*), leftmargin=1.4em]
\item \textbf{Statistical dependence} Two variables $X$ and $Y$ are (marginally or conditionally) \emph{dependent} if their joint distribution under passive observation does not factorize, e.g., $X\!\not\!\perp\!\!\!\perp Y\mid Z$. Dependence is a purely observational property.
\item \textbf{Causal relationship} We say $X$ is a \emph{cause} of $Y$ (denoted $X\!\to\!Y$ in the causal graph) when intervening on $X$ changes the distribution of $Y$: $P(Y\,|\,\operatorname{do}(X=x))$ depends on $x$ \cite{pearl2009causality}. Causation implies a directed edge in the underlying \emph{interventional} DAG, but \underline{does not} require $X$ and $Y$ to be statistically dependent (collider structures are the standard counter-example).
\end{enumerate}

The \textit{Causal Markov Condition (CM)} states that, given the causal graph \( \mathcal{G} \), the joint probability distribution \( P(\mathbf{X}) \) factorizes as:
\begin{equation}\label{eq:factorization}
    P(X_1, X_2, \ldots, X_n) = \prod_{i=1}^{p} P(X_i | \text{Pa}(i, \mathcal{G})),
\end{equation}
where \( \text{Pa}(i, \mathcal{G}) \) represents the parents of \( X_i \) in the graph. In other words, the CM states that, given a causal graph, each variable is conditionally independent of its non-descendants, given its parents in the graph. This condition is crucial for causal inference as it ensures that the relationships in the graph can be used to explain the dependencies in the data.

\begin{definition}[Causal Markov property]\label{def:markov-property}
\(P\) is \emph{Markov with respect to} \(\mathcal{G}\) if each variable is independent of its non-descendants, conditioned on its parents:
\[
X_i\!\perp\!\!\!\perp\! \operatorname{nd}(X_i)\,\big|\,\operatorname{Pa}(X_i), \qquad \forall\,i\in\{1,\ldots,p\}.
\]
\end{definition}

\textit{Faithfulness} refers to the assumption that all conditional independencies present in the observed data are reflected in the structure of the causal graph. Formally, the distribution $P$ in Eq. \eqref{eq:factorization} is said to be faithful to $\mathcal{G}$ if the only conditional independencies in $P$ are those implied by the d-separation in $\mathcal{G}$ \cite{spirtes2000causation, pearl2009causality}. This assumption ensures that there are no accidental independencies in the data due to parameter values (e.g., when causal effects perfectly cancel each other out).

\begin{definition}[Faithfulness]\label{def:faithfullness}
\(P\) is \emph{faithful to} \(G\) if every conditional independence that holds in \(P\) is \emph{entailed} by the DAG via \(d\)-separation:
\[
A\perp\!\!\!\perp B \,\big|\,C \text{ in } P \;\Longrightarrow\; A\;\text{and}\;B\;\text{are \(d\)-separated by}\;C\text{ in }G, \quad \forall\,A,B,C\subseteq V.
\]
Equivalently, the set of conditional independences of \(P\) coincides with the set implied by \(G\).
\end{definition}

Faithfulness, combined with the Causal Markov Condition, is essential for inferring the correct causal structure from observational data, as it guarantees that the observed statistical relationships correspond directly to the structure of the causal graph. Additionally, in the causal discovery setting, the underlying model that generates the data is assumed to be a general Structural Equation Model (SEM) as follows:
\begin{equation}\label{eq:sem}
    X_i = \f_i ( X_{\text{Pa}(i, \mathcal{G})}, \varepsilon_i ), \ \text{with} \ \varepsilon_i \sim \mathcal{N}(0, 1),\ \text{for}\ i = 1, \ldots, p,
\end{equation}
where $ \f_i $ is a function from $\mathbb{R}^{|\text{Pa}(i, \mathcal{G})|+1} \rightarrow \mathbb{R}$, and $\varepsilon_i$ is a unit centered Gaussian noise. SEM is a framework for modeling causal relationships where each variable is expressed as a function of its direct causes (parents) and an error term, typically assumed to be independent noise. SEMs are also essential for causal discovery because they provide a formal way to represent and quantify the effects of interventions and establish the relationships between variables.

Causal discovery aims to infer cause-effect relationships between variables in observational data, typically represented by a causal graph. A causal graph, often modeled as a DAG, consists of nodes representing variables and directed edges indicating causal influences. The primary challenge in causal discovery then, is to identify these causal relationships without direct evidence from interventions, relying instead on specific assumptions about the data and its underlying generative processes.

\begin{definition}[Causal-discovery problem]
Given \(m\) i.i.d.\ samples from an \emph{unknown} \(P\) that is Markov and faithful to some (unknown) DAG \(\mathcal{G}\), \emph{causal discovery} asks for an estimator that, as \(n\to\infty\), identifies with high probability the \emph{Markov-equivalence class}
\begin{equation}    
\mathcal{M}(P)\;=\; \bigl\{\,\mathcal{G} \;\big|\; P\;\text{is Markov and faithful to }\mathcal{G} \bigr\},
\end{equation}
or a representative DAG in that class.
\end{definition}

Throughout the paper we work under these standard assumptions (\cite{pearl2009causality, spirtes2000causation}); Section \ref{s:Results} empirically evaluates how the proposed \methodname algorithm performs when the assumptions are met or mildly violated.

While causal discovery \emph{leverages} patterns of statistical (in)dependence under the Markov and Faithfulness assumptions, the terms \textit{causal relation} and \textit{statistical dependence} are not interchangeable. The synthetic experiments in Appendix~A (confounder, chain, collider) illustrate cases where high marginal dependence hides a non-causal variable and vice-versa.

\subsection{SHAP values and causal relationships} \label{s:Shap-and-causal}

When the predictor is sufficiently expressive, adding a variable \(X_j\) to an existing feature set \(S\) improves the out-of-sample prediction of the target \(Y\) \emph{only if} \(X_j\) contributes information about \(Y\) that is \emph{not} already contained in \(S\).  Hence, a near-zero marginal improvement suggests the conditional independence \(X_j \perp\!\!\!\perp Y \mid S\).  This link between lack of predictive gain and conditional independence underlies many \emph{causal-discovery} algorithms, which use such independencies to distinguish direct causal links from indirect (mediated) or spurious (confounded) associations when building a causal graph.

\paragraph{Related SHAP--causality studies}
Several recent papers have examined how Shapley attributions might interface with causal reasoning \cite{heskes_causal_2020}. introduces \emph{causal SHAP}, which \emph{assumes} a known DAG and re-weights coalitions accordingly; \cite{min_interpretability_2023} applies standard SHAP values to a domain-specific regression task (coal-bed-methane wells) and discuss causal plausibility qualitatively; and \cite{xu_causality_2020} surveys ways in which explainers may hint at causality, but stops short of a data-driven discovery algorithm. In contrast, the present work derives an explicit link between aggregated SHAP values and conditional dependence (Section~\ref{ss:mathematical-foundation}, Eq.~\ref{eq:approximation}) and embeds that link in a full pipeline (Sections~\ref{s:proposed-method}--\ref{s:Results}) that \emph{discovers} the causal graph directly from observational data without prior causal knowledge.

\subsection{Mathematical Foundation} \label{ss:mathematical-foundation}

In this section, we establish the theoretical connection between Shapley values and conditional independence within the context of causal discovery. This foundation supports the use of Shapley values in identifying causal relationships by relating feature importance measures to probabilistic dependence structures.

Let $F$ be the set of all features (variables), excluding the target variable \( X_i \). Each feature \( X_j \in F\) is considered a ``player'' in a cooperative game where the objective is to predict \( X_i \). 
We define a function \( \hat{f}(S): 2^{|F|} \rightarrow \mathbb{R} \) that maps a subset of features \( S \subseteq F \) to a real number representing the contribution of \( S \) to predicting \( X_i \), with $|F|$ being the cardinality of $F$. The classical definition of the Shapley value \( \phi_j \) for feature \( X_j \) is given by:
\begin{equation}
\label{eq:shap-values}
\phi_j = \sum_{S \subseteq F_{\setminus \{X_j\}}} \frac{|S|! \, (|F| - |S| - 1)!}{|F|!} \left[ \hat{f}(S \cup \{X_j\}) - \hat{f}(S) \right],
\end{equation}
where \( \hat{f}(S \cup \{X_j\}) - \hat{f}(S) \) represents the marginal contribution of feature \( X_j \) to the prediction of \( X_i \) given subset \( S \).

\subsubsection{Marginal Contributions and Conditional Independence}

In \cite{ma_predictive_2020}, a connection between Shapley value summands and conditional independence is demonstrated by relating conditional independence in a faithful Bayesian network with the summands of the Shapley value. Specifically, within this framework, if a variable \( X_j \) provides additional information about the target variable given a subset \( S \) of other variables, this relationship is captured by the Shapley value summands for \( X_j \) given \( S \). This interpretation aligns with the concept of conditional independence, where the Shapley summand reflects the additional predictive contribution of \( X_j \) in the context of \( S \).  In causal discovery, assuming causal sufficiency and faithfulness, SHAP values can indicate conditional independence relationships within a causal DAG, providing an interpretable approach to uncovering causal relationships rather than mere associations.

To connect Shapley values with conditional independence, consider the marginal contribution \( \Delta_{j,S} \) of feature \( X_j \) for a specific subset \( S \subseteq F_{\setminus \{X_j\}} \), given by $\Delta_{j,S} = \hat{f}(S \cup \{X_j\}) - \hat{f}(S)$. Assuming that \( \hat{f}(S) \) accurately reflects the predictive power of \( S \) for \( X_i \), the marginal contribution \( \Delta_{j,S} \) relates to the conditional dependence between \( X_i \) and \( X_j \) given \( S \). Specifically:

\begin{itemize}
    \item If \( X_j \) provides no additional information about \( X_i \) given \( S \), then \( X_i \) is conditionally independent of \( X_j \) given \( S \), denoted as \( X_i \ci X_j \mid S \). In this case $p(X_i \mid X_j, S) = p(X_i \mid S)$, and the marginal contribution is zero ($\Delta_{j,S} = 0$).
    \item If \( X_j \) provides additional predictive information about \( X_i \) given \( S \), then \( X_i \) is conditionally dependent on \( X_j \) given \( S \), denoted as \( X_i \not\!\perp\!\!\!\perp X_j \mid S \). In this case:
    \begin{equation}
    \label{eq:conditional-dependence}
    p(X_i \mid X_j, S) \ne p(X_i \mid S),
    \end{equation}
    and the marginal contribution is positive ($\Delta_{j,S} > 0$).
\end{itemize}

\subsubsection{Aggregating over all subsets}

The Shapley value $\phi_j$ aggregates the marginal contributions $\Delta_{j,S}$ over all subsets $S$ of $F_{ \backslash\{X_j\}}$, weighted by the Shapley weights:
\begin{equation} \label{eq:shapley-weight}
w(S) = \frac{|S|! \, (|F| - |S| - 1)!}{|F|!}.
\end{equation}

These weights satisfy \( \sum_{S} w(S) = 1 \) and reflect the importance of each subset in the calculation.

\begin{remark}
Equations\,(\ref{eq:shap-values}) and (\ref{eq:shapley-weight}) are reproduced verbatim from the original work of Shapley (1953)\,\cite{shapley1953value}; we include them only to fix notation.
\end{remark}

\subsubsection{Shapley-weighted Conditional Dependence Indicator}

To formalize the relationship between Shapley values and conditional independence, we introduce a new indicator function \( I_{j,S} \):
\begin{equation}
I_{j,S} =
\begin{cases}
1, & \text{if } p(X_i \mid X_j, S) \ne p(X_i \mid S), \\
0, & \text{if } p(X_i \mid X_j, S) = p(X_i \mid S).
\end{cases}
\end{equation}

Using this indicator function, the Shapley value $\phi_j$ can be expressed as:
\begin{equation}
\phi_j = \sum_{S} w(S) \cdot \Delta_{j,S} = \sum_{S} w(S) \cdot I_{j,S} \cdot \Delta_{j,S}.
\end{equation}

Assuming that the average marginal contribution $\overline{\Delta_j}$ is approximately constant across subsets where $X_j$ is conditionally dependent on $X_i$, we can approximate $\phi_j$ as:
\begin{equation}
\phi_j \approx \overline{\Delta_j} \cdot \sum_{S} w(S) \cdot I_{j,S}.
\end{equation}

\subsubsection{Weighted probability of conditional dependence}

The term \( \sum_{S} w(S) \cdot I_{j,S} \) represents the weighted probability that \( X_j \) is conditionally dependent on \( X_i \) across all subsets \( S \), under the probability distribution defined by the Shapley weights \( w(S) \), i.e.,:
\begin{equation}
P_{\text{weighted}}(X_j \text{ is conditionally dependent on } X_i) = \sum_{S} w(S) \cdot I_{j,S}.
\end{equation}

Thus, the Shapley value \( \phi_j \) can be interpreted as:
\begin{equation} \label{eq:approximation}
\phi_j \approx \overline{\Delta_j} \cdot P_{\text{weighted}}(X_j \text{ is conditionally dependent on } X_i),
\end{equation}

where $P_{\text{weighted}}(\cdot)$ abbreviates the Shapley weighting over coalitions. Equation \eqref{eq:approximation} states that a feature's Shapley value $\phi_j$ is proportional to both the average dependence strength $\Delta_j$ and the Shapley-weighted probability that $X_j$ is conditionally dependent on $Y$. \cite{ma_predictive_2020}
showed that a \emph{single} Shapley summand $f(S\cup\{X_j\})-f(S)$ vanishes if and only if $X_j\perp\!\!\!\perp Y\,|\,S$ in a faithful Bayesian network. In contrast, Eqs.\,(6)--(10) aggregate \emph{all} summands under the original Shapley weights, yielding the \emph{Shapley-weighted probability of conditional dependence} in Eq.\,(10)—a quantity that did not appear in \cite{ma_predictive_2020} and underpins the causal-discovery criterion of the present work.

Controlled synthetic experiments (Appendix~\ref{app:shapley-validation}, Table~\ref{tab:aggregated-results}) support this claim across confounder, chain, collider, and collinear structures (see Section~\ref{ss:divergence-between-dependence-and-shap}).

\subsubsection{Implications for causal discovery} \label{sss:implications}

The relationship between Shapley values and conditional dependence suggests that features with higher Shapley values are more likely to be conditionally dependent on the target variable \( X_i \), while features with lower Shapley values are more likely to be conditionally independent. This connection provides a theoretical justification for using Shapley values in causal discovery. 

More formally, a high Shapley Value (\( \phi_j \) large) indicates that \( X_j \) frequently contributes significant predictive information about \( X_i \) across various subsets \( S \), suggesting a potential causal relationship or strong association. On the contrary, a low Shapley Value (\( \phi_j \) small) implies that \( X_j \) rarely provides additional predictive information about \( X_i \), indicating possible conditional independence and a lower likelihood of a direct causal link.

Leveraging this theoretical foundation, Shapley values can be utilized to estimate the conditional dependence structure among variables. However, it is important to acknowledge the assumptions underlying this connection, particularly for the approximation in Eq.~\eqref{eq:approximation}. This approximation presumes that the average marginal contribution \(\overline{\Delta_j}\) is relatively consistent across subsets where \(X_j\) is conditionally dependent on \(X_i\), though in practice, individual marginal contributions \(\Delta_{j,S}\) can vary significantly, making the constant \(\overline{\Delta_j}\) a simplification. Furthermore, the predictive function \(\hat{f}(S)\) is assumed to accurately capture true conditional probabilities $p(X_i | S)$. Finally, for a causal interpretation of the SHAP-derived insights within a discovery context, the core assumptions of faithfulness and causal sufficiency are necessary to ensure that inferred relationships represent causal connections rather than mere statistical dependencies.

Despite these assumptions, the established relationship provides valuable insights into how Shapley values can reflect the underlying conditional independence structures, supporting their use in causal inference tasks.

\subsubsection{Divergence between dependence and Shapley} \label{ss:divergence-between-dependence-and-shap}

The theoretical connection articulated above, particularly Equation \eqref{eq:approximation}, is supported by controlled synthetic experiments (see Appendix \ref{app:shapley-validation}). These experiments, conducted across canonical causal structures (confounders, chains, colliders, and collinear parents), highlight several key insights:

\noindent \textbf{Divergence from single-set conditional independence (CI) tests}. Shapley values differ notably from traditional single-set CI tests (e.g., Fisher-Z tests). For example, in the confounder ($Z \rightarrow X$, $Z \rightarrow Y$) and chain ($X \rightarrow Z \rightarrow Y$) structures predicting $Y$, a feature $X$ may appear conditionally independent of $Y$ given $Z$ (yielding a non-significant Fisher-Z p-value, i.e., $X \ci Y \mid Z$), yet still receive substantial non-zero Shapley values. This occurs because Shapley values aggregate contributions across multiple feature coalitions, including subsets not conditioning on $Z$, thereby capturing broader predictive relevance.

\noindent \textbf{Robustness under high correlation scenarios}: In scenarios involving highly collinear predictors (e.g., $X\_1 \approx X\_2 \rightarrow Y$), Shapley values exhibit greater stability and interpretability compared to traditional CI tests or regression coefficients, which often become unstable or ambiguous. Empirical results consistently show stable Shapley values for collinear features, reflecting intuitive allocation of feature importance, consistent with properties of interventional explainers like TreeExplainer (discussed in Section \ref{ss:Multicollinearity}).

\noindent \textbf{Sensitivity to true causal drivers:} Features that act as direct causes or strong mediators consistently receive high Shapley values, accurately reflecting high conditional dependence probabilities. For instance, mediators and direct causal features in confounder, chain, or collider structures consistently display elevated Shapley values, highlighting their predictive importance.

Overall, these empirical findings underscore that Shapley values, despite being model-dependent, provide comprehensive and robust assessments of feature contributions by aggregating across all feature subsets, making them valuable for causal discovery methods like \methodname under assumptions of faithfulness and causal sufficiency.


\section{Proposed method} \label{s:proposed-method}
 
The proposed method \methodname approaches the causal discovery problem in a series of steps (see Fig.~\ref{fig:diagram}). First, we fine-tune and train a series of regressors to predict each variable $X_i$, for $i=1, \ldots, p$, using the remaining \(p-1\) variables in the dataset, denoted as $X_{\backslash i} = \{X_1, \ldots, X_{i-1}, X_{i+1}, \ldots, X_p\}$. Once the regressors are fine-tuned and trained, we compute Shapley values to estimate the contribution of each variable to the prediction, applying a bootstrapping mechanism, i.e., iteratively using different samples of the dataset. Through this repeated sampling, \methodname identifies a robust set of features that can be considered the potential causes (i.e., $\text{Pa}(X_i)$) for each target variable $X_i$. As a final step, results from the different regressors are combined to obtain a plausible causal graph, which is reviewed to  direct edges and remove eventual cycles. Next, we describe these steps in detail. It is worth noting that while the overall structure of the \methodname pipeline—training predictive models, assessing feature contributions, and constructing a graph—is a general approach, the core novelty of \methodname lies in its specific instantiation using Shapley values for robust feature impact assessment and parent selection, guided by the theoretical connections discussed in Section~\ref{ss:mathematical-foundation}, and its subsequent steps for graph refinement.

\begin{figure}[ht!]
    \centering
    \resizebox{0.75\columnwidth}{!} {
        \begin{tikzpicture}[
            node distance=0.5cm,
            every node/.style={rectangle, draw, fill=white, rounded corners, align=center, minimum width=1cm, minimum height=1cm},
            arrow/.style={->, thick},
            loopback/.style={->, thick, dashed, bend left=45},
        ]
        
        \node (start) [draw=none, fill=none] {Start};
        \node (train) [right=of start] {Train regressor \\ models};
        \node (sample) [right=of train] {Sample data};
        \node (shap) [right=of sample] {Compute SHAP \\ Values ($\Phi$)};
        \node (select) [right=of shap] {Select parents for \\ each feature};
        \node (dag) [right=of select] {Update Adjacency \\ Matrix};
        \node (direct) [below=of sample] {Direct Edges};
        \node (combine) [right=of direct] {Combine DAGs \\ from each regressor};
        \node (cycles) [right=of combine] {Remove Cycles};
        \node (end) [right=of cycles, draw=none, fill=none] {Final DAG};
        
        \coordinate[above=of sample] (x);
        \draw [dashed, ->] (dag.north) -- ++(0, 0.3cm) |- (x) -- (sample.north);
        \node (bootstrap) [draw=none] at (9, 1) {Bootstrapping};
        
        \node (shap) [right=of sample] {Compute SHAP \\ Values ($\Phi$)};
        \node (select) [right=of shap] {Select parents for \\ each feature};
        
        \draw [arrow] (start) -- (train);
        \draw [arrow] (train) -- (sample);
        \draw [arrow] (sample) -- (shap);
        \draw [arrow] (shap) -- (select);
        \draw [arrow] (select) -- (dag);
        \draw [arrow] (dag.east) -- ++(0.3cm, 0) |- (3, -0.75) |- (direct.west); 
        \draw [arrow] (direct) -- (combine);
        \draw [arrow] (combine) -- (cycles);
        \draw [arrow] (cycles) -- (end);
        
        \end{tikzpicture}
    }

    \caption{Overview of the \methodname workflow. The process begins with training regressors (Section~\ref{s:Models-training}), followed by a bootstrapping procedure (Algorithm~\ref{alg:dag-construction}) that includes data sampling, computation of SHAP values ($\phi$) to assess feature impact (Section~\ref{ss:computing-shap-values-for-feature-impact}), selection of candidate parents based on these values (Algorithm~\ref{alg:parent-selection}, Section~\ref{ss:parents-selection}), and updating the adjacency matrix (Section~\ref{ss:adjacency-matrix-update}). Edges are then oriented (Section~\ref{ss:directing-edges}), and outputs from multiple regressors are combined and pruned to produce the final DAG (Section~\ref{ss:final-dag}). While parts of the pipeline are general, \methodname is specifically designed to leverage Shapley values for assessing feature contributions, as motivated in Section~\ref{ss:mathematical-foundation}.}
    \label{fig:diagram}
\end{figure}

\subsection{Models training} \label{s:Models-training}

The first step in deriving the causal model is to train a separate model to predict each variable in the dataset, with hyper-parameter optimization (HPO). Before model training, the data is normalized by removing the mean and scaling to unit variance. To ensure robustness, we employ two complementary regressors: a deep feed-forward neural network (DFN) and gradient boosting trees (GBT). DFNs are highly flexible models capable of approximating any function, leveraging their capacity as universal approximators, as demonstrated by \cite{hornik1989multilayer}. GBTs, known for their effectiveness in handling tabular continuous data and offering more interpretable outputs as shown in \cite{friedman2001greedy}, serve to validate and complement the results obtained from the DFN models, with the XGBoost implementation used in this study. Although the \methodname framework allows for the integration of additional regressors, the results from DFN and GBT have proven to be sufficiently robust for our purposes (see Appendix~\ref{app:single-vs-union}).

A total of $p$ DFN and $p$ GBT models are trained to predict each feature $X_i$ ($i=1, \ldots, p$) from the remaining variables $X_{\backslash i}$, using mean squared error (MSE) as the loss function. To enable DFNs, which are deterministic architectures, to effectively model stochastic relationships inherent in the data-generating process (Eq.~\ref{eq:sem}), we incorporate an additional i.i.d.\ Gaussian noise variable $\nu \sim \mathcal{N}(0, \sigma^2)$ as an input feature. This technique, inspired by generative causal modeling \cite{kalainathanGenerativeNeuralNetworks2020} and with $\nu$ re-generated at each training step, allows the DFN to use this input noise to model the influence of the exogenous SEM noise term $\varepsilon_i$, rendering the learned predictive functions $\hat{f}_{i}(X_{\setminus i}, \nu)$ inherently non-deterministic. This imparted stochasticity is crucial for \methodname as it: (i) yields more stable and graduated Shapley scores, particularly with low-noise or deterministic synthetic data, by preventing numerical artifacts that can arise from perfectly deterministic learned functions; and (ii) encourages DFNs to capture richer conditional distribution information beyond just the mean, thus leading to a more nuanced SHAP-based feature importance assessment for robust parent selection (Algorithm~\ref{alg:parent-selection}). The trained models are therefore defined as $\hat{f}_{i}(X_{\backslash i}, \nu) = \hat{X_i} \approx X_i$ for DFNs, and $\hat{f}_{i}(X_{\backslash i}) = \hat{X_i} \approx X_i$ for GBTs (without the additional noise input).

The model hyperparameters are determined in the training phase for each dataset, using a tree-structured Parzen estimator (TPE) \cite{bergstra2011algorithms} with an 80/20 split for train and validation. Hyperparameter optimization is a critical step, as one of the assumptions stated in Section \ref{sss:implications} is that regressor \normalsize $\hat{f}$ accurately captures the true conditional probabilities.

\subsection{Bootstrapping and initial causal graph construction}

In this phase, we implement the bootstrapping approach to construct an initial undirected causal graph ($\widehat G_{\text{undir}}$). This process involves three main steps: computing SHAP values to assess feature importance, selecting potential parent features based on their impact, and updating an adjacency matrix that represents candidate causal connections. These steps are iteratively repeated in the bootstrapping loop (Algorithm \ref{alg:dag-construction}) to build a stable causal structure. 

\begin{algorithm}[htb]
    \footnotesize
    \caption{Bootstrapped DAG Construction}
    \label{alg:dag-construction}
    \begin{algorithmic}[1]
    \Require Let $\hat{f}$ be a fine-tuned regressor model (DFN or GBT)
    \State Initialize empty adjacency matrix $\mathbf{A} \in \mathbb{R}^{p \times p}$
    \For{$t = 1$ to $T$} \Comment{Number of bootstrap iterations}
        \State Sample a subset $\mathbf{X}^{(t)}$ from $\mathbf{X}$ (observational data)
        \For{each feature $X_i$}
            \State Fit SHAP explainer with $\mathbf{X}^{(t)}_{\smallsetminus i}$ and model $\hat{f}$ to obtain \ $\Phi^{(i)} = \{\phi_1, \ldots, \phi_{i-1}, \phi_{i+1}, \ldots, \phi_p\}$
            \State Run Algorithm~\ref{alg:parent-selection} with $\Phi^{(i)}$ to obtain $\textit{Pa}(X_i)$
            \For{each $X_j \in \textit{Pa}(X_i)$}
                \State Increment the corresponding entry: $a_{i,j} \gets a_{i,j} + 1$
            \EndFor
        \EndFor
    \EndFor
    \State Normalize: $a_{i,j} \gets \tfrac{a_{i,j}}{T}$ for all $(i,j)$
    \State Filter edges: $a_{i,j} \gets 0$ if $a_{i,j} < \tau$
    \State \Return Adjacency matrix $\mathbf{A}$ representing a \emph{stable} undirected graph $\widehat G_{\text{undir}}$
    \end{algorithmic}
\end{algorithm}

\begin{algorithm}[htb]
    \footnotesize
    \caption{Parent Selection from SHAP Values}
    \label{alg:parent-selection}
    \begin{algorithmic}[1]
    \Require Input array $\Phi = \{\phi_1, \ldots, \phi_{p-1}\}$ (SHAP values)
    \State $\Lambda \gets$ Pairwise Euclidean distances in $\Phi$
    \State $\Lambda' \gets$ Sort elements of $\Lambda$ in descending order
    \State Initialize $\zeta \gets \max(\Lambda) + \text{constant}$
    \Repeat
        \State Run DBSCAN on $\Phi$ with parameter $\zeta$
        \State $n \gets$ Number of clusters
        \If{$n = 1$}
            \State $\zeta \gets \zeta - \max(\Lambda')$
            \State Remove $\max(\Lambda')$ from $\Lambda'$
        \EndIf
    \Until{$n > 1$ or $\Lambda' = \emptyset$}
    \If{no clusters formed}
        \Return \texttt{None}
    \Else
        \Return Features in cluster with highest mean SHAP value
    \EndIf
    \end{algorithmic}
\end{algorithm}

\subsubsection{Computing SHAP values for feature impact} \label{ss:computing-shap-values-for-feature-impact}

The first step in Algorithm \ref{alg:dag-construction}, after sampling a subset of the data, is to compute SHAP values for each feature to identify its impact on the target variable in each regressor model. For a given model (DFN or GBT), the SHAP value \( \phi_j \) is calculated for each feature \( X_j \) when predicting the target variable \( X_i \). This value indicates the marginal contribution of \( X_j \) to the prediction of \( X_i \), allowing us to interpret \( \phi_j > 0 \) as evidence that \( X_j \) is likely to influence \( X_i \) (i.e., \( X_i \nci X_j \)). This computation provides a foundation for identifying variables that may play a causal role in influencing each target variable, as described in Section \ref{sss:implications}.

\subsubsection{Parents selection} \label{ss:parents-selection}

In the second step of Algorithm \ref{alg:dag-construction}, we apply the DBSCAN clustering algorithm to group features based on their SHAP values and select those that have a significant impact on predicting the target variable. Feature clusters are adjusted dynamically by decreasing the neighborhood radius parameter \( \zeta \) until more than one group are detected, helping to separate influential features from less impactful ones. This clustering process effectively selects potential parent features, as it isolates variables that likely have causal influence on the target. Algorithm \ref{alg:parent-selection} outlines this clustering procedure in detail, and Appendix~\ref{app:feature-selection} describes the process with a visual example and additional details. 

DBSCAN is selected for clustering SHAP values because, unlike methods such as K-Means, it does not require specifying the number of clusters upfront—a crucial advantage given the data-dependent and unknown grouping of feature importances. Moreover, DBSCAN intrinsically distinguishes influential parent candidates (high SHAP values) from negligible-impact features by labeling low-density points as noise, a capability partition-based methods lack. Algorithm~2 further improves DBSCAN by adaptively setting the neighborhood radius parameter $\zeta$ according to the empirical distribution of SHAP values, avoiding arbitrary thresholds and providing a principled distinction between \textit{salient} and \textit{negligible} feature contributions.

Simpler heuristic methods, such as using average SHAP values with percentile thresholds, were explored but produced worse results, as they struggled to effectively distinguish between features with varying levels of impact.

\subsubsection{Adjacency matrix update and graph construction} \label{ss:adjacency-matrix-update}

With the influential features identified, we proceed to construct an initial causal graph by updating an adjacency matrix, which records candidate edges between features. In each bootstrapping iteration, a subset of the data \( \mathbf{X}^{(t)} \) is sampled, and SHAP-based parent selection is performed. For each pair of selected features, the corresponding entry in the adjacency matrix \( \mathbf{A} \) is incremented, capturing the frequency of selection across iterations. After bootstrapping, we apply a threshold \( \tau \) to filter out edges with low frequencies, retaining only stable and meaningful connections. This thresholding process yields a plausible undirected causal graph, representing the initial structure without directed edges ($\widehat G_{\text{undir}}$). Thus, the stability of an edge in this initial graph is directly determined by the parameter $\tau$; an edge is included if its frequency of selection during the bootstrapping procedure (Algorithm~\ref{alg:dag-construction}) meets or exceeds this threshold, signifying its robustness against variations in data subsamples.

Step~3 of Algorithm~\ref{alg:dag-construction} involves creating a bootstrap sample $X^{(t)}$ by selecting a subset of the original data samples (i.e., rows) from the full dataset $X$. To ensure robust coverage of the dataset across the $T$ bootstrap iterations, the proportion $c$ of original data samples to be included in each such bootstrap sample $X^{(r)}$ is set using the formula $c \ge 1 - q^{1/T}$. In this formula, $q$ is the maximum acceptable probability that any given original data sample (row) is not selected across any of the $T$ bootstrap iterations (typically $q=0.01$), and $T$ is the total number of iterations for the bootstrapping mechanism (typically $T=50$, as specified for Algorithm~\ref{alg:dag-construction}).

The undirected graph $\widehat G_{\text{undir}}$ produced at this stage is therefore stable under bootstrap perturbations; only these vetted edges are passed to the subsequent orientation module of Section~3.3.

\subsection{Directing edges} \label{ss:directing-edges}

Once a \emph{stable} undirected graph $\widehat G_{\text{undir}}$ has been obtained—\textit{i.e.}, after the bootstrapping procedure of Section~\ref{ss:adjacency-matrix-update} has normalized the adjacency matrix and applied the frequency threshold~$\tau$ in Algorithm~1 (lines~12--13) to retain only edges with selection probability $f_{ij}\ge\tau$—the next step is to orient the remaining candidate edges. The direction of each such pre-identified edge $(X_i, X_j)$ is established using principles from Additive Noise Models (ANMs), as described in \cite{PetersJanzingSchoelkopf17}. This ANM-based orientation procedure involves fitting regression models for both potential directions (e.g., $X_j$ as a function of $X_i$, and $X_i$ as a function of $X_j$).

To determine the causal direction, we then use the Hilbert-Schmidt Independence Criterion (HSIC) \cite{gretton2007kernel} to test for independence between the residuals of each regression and the respective predictor variable. For a potential edge $X_i \rightarrow X_j$, if the HSIC test accepts the hypothesis that the residuals from regressing $X_j$ on $X_i$ are independent of $X_i$, we infer the causal direction $X_i \rightarrow X_j$. Conversely, if the HSIC test indicates \textit{dependence} between the residuals and $X_i$ (i.e., the null hypothesis of independence is rejected, typically if the test $p$-value is below a significance threshold), this direction $X_i \rightarrow X_j$ is considered not supported by the ANM assumption. The process is repeated for the reverse direction. Thus, the HSIC test serves to ascertain whether the statistical dependence flagged by the Shapley values aligns with an identifiable ANM structure in one direction over the other.

The resulting graph is a directed acyclic graph (DAG) that captures the inferred causal relationships between the features.

\subsection{Final DAG} \label{ss:final-dag}

The methodology presented here culminates in the generation of the final DAG, denoted as $\mathcal{G}_{\text{\tiny{\methodname}}}$. This final DAG is produced by taking the union ($\cup$) of the two DAGs $\mathcal{G}_{\text{\tiny{DFN}}}$ and $\mathcal{G}_{\text{\tiny{GBT}}}$, which are generated by the described procedure when using DFN and GBT as the regressors, respectively (see Appendix~{\ref{app:single-vs-union}} for a performance comparison among $\mathcal{G}_{\text{\tiny{\methodname}}}$, $\mathcal{G}_{\text{\tiny{DFN}}}$ and $\mathcal{G}_{\text{\tiny{GBT}}}$).

The union ($\cup$) is chosen to leverage the complementary strengths of these diverse model types in identifying potential causal links, aiming for a more comprehensive discovery of true edges. This strategy is empirically supported by a higher overall F1-score compared to using intersection or individual regressors, as detailed in Appendix \ref{app:single-vs-union}. The implementation provides a configurable parameter that allows users to choose between computing the union or the intersection, depending on their specific requirements.

The final DAG $\mathcal{G}_{\text{\tiny{\methodname}}}$ must be acyclic to represent valid causal relationships. Hence, in those cases where the new proposed DAG contains cycles or bidirectional edges between two nodes, as a result of the union of two DAGs, we employ a strategy based on the use of the \emph{SHAP discrepancy} $\delta_{j}^{(i)}$ to make it acyclic.

The SHAP discrepancy is introduced as a measure to assess how well the Shapley values of a feature explain the variability in the target variable. It quantifies the extent to which the contributions of a feature \( X_j \) (captured by its Shapley values \( \phi_j \)) align with the actual values of \( X_i \) (see Appendix \ref{app:shap-discrepancy} for more details), computed as follows:
\begin{equation}
\label{eq:shap-discrepancy}
    \delta_{j}^{(i)} = 1 - R^2(X_i, \phi_j) = \frac{\sum_{k=1}^{m} (x_{k,i} - \phi_{k,j})^2}{\sum_{k=1}^{m} (x_{k,i} - \overline{X_i})^2},
\end{equation}
where \( x_{k,i} \) is the \( k \)-th observation of \( X_i \), \( \phi_{k,j} \) is the corresponding Shapley value of \( X_j \) for predicting \( X_i \), \( \overline{X_i} \) is the mean of \( X_i \), and \( m \) is the total number of samples.

A lower SHAP discrepancy indicates that the Shapley values \( \phi_j \) closely approximate \( X_i \), suggesting a stronger potential causal influence from \( X_j \) to \( X_i \). In the context of removing cycles in the DAG, the SHAP discrepancy helps identifying which edges are less supported by the data. Briefly, when a cycle is detected, edges with higher discrepancies are considered for reorientation or removal to break the cycle, ensuring the resulting graph accurately represents the underlying causal relationships.

The proposed method iteratively examines each detected cycle and seeks to resolve it by removing edges with the highest SHAP discrepancy. In particular, for each cycle, the proposed strategy computes the SHAP discrepancies for each edge in the cycle. By comparing these discrepancies, it identifies edges that could reduce the overall discrepancy by reversing their direction, indicating potential misorientations. Reorienting such edges can break the cycle while preserving the causal relationships suggested by the data. If reorientation is insufficient or not possible, the proposed method removes the edge with the largest SHAP discrepancy—implying it is the weakest link in the cycle—to eliminate the cyclicity.

This process ensures that the final graph is acyclic and retains the most significant causal structures inferred from the data. The algorithm effectively balances the need to break cycles with the goal of maintaining the integrity of the underlying causal relationships.

\subsection{Visual example}

\begin{figure*}[ht]
    \centering
    \includegraphics[width=0.65\linewidth]{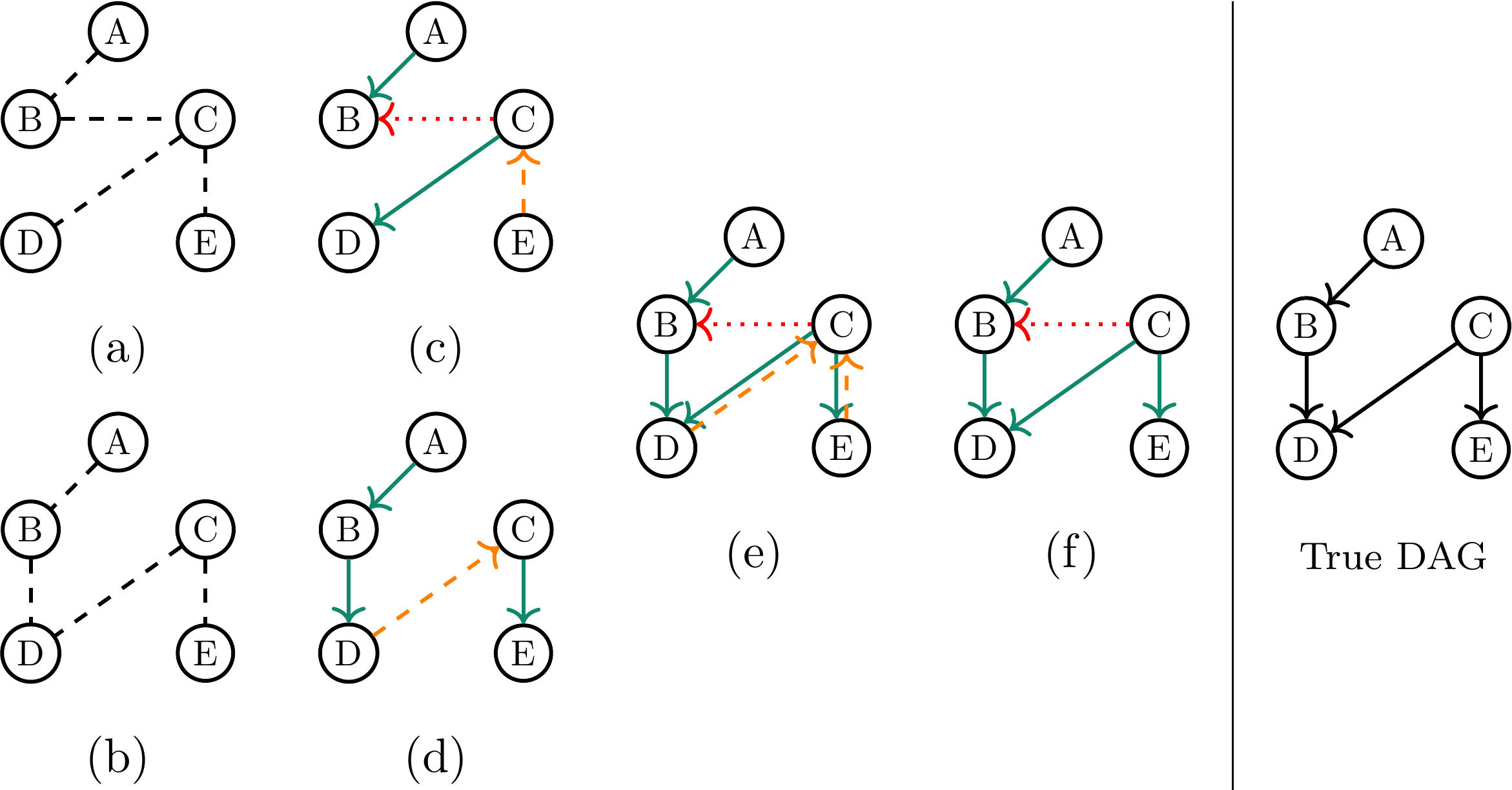}
    \captionof{figure}{A visual summary 
    of the intermediate graphs  generated by \methodname to obtain the final DAG $\mathcal{G}_{\text{\tiny{\methodname}}}$. (a)-(b) represent the unoriented graphs for each of the two regressors after running the steps described in Sections \ref{s:Models-training}, \ref{ss:computing-shap-values-for-feature-impact}, \ref{ss:parents-selection} and \ref{ss:adjacency-matrix-update}. (c)-(d) are the result of establishing the direction of each edge presented in (a)-(b), respectively, following the approach described in Section \ref{ss:directing-edges}. (e) is the initial DAG resulting from the union of DAGs (c)-(d). (f) is final DAG $\mathcal{G}_{\text{\tiny{\methodname}}}$ resulting from removing cycles and/or bidirectional edged from (e), following the steps described in Section \ref{ss:final-dag}. The true DAG is also depicted for comparison. Green arrows represent correctly predicted edges, orange ones edges with the incorrect direction, and red ones edges that are missing in the true DAG.}
    \label{fig:summary}
\end{figure*}

The intermediate DAGs obtained by \methodname are summarized, in a simplified way, in Fig.~\ref{fig:summary}. Each of the regressors tuned and trained to predict each of the variables in the dataset (Section \ref{s:Models-training}) are used to compute SHAP values (Section \ref{ss:computing-shap-values-for-feature-impact}), and from them, select the candidate parents (causes) of each variable (Section \ref{ss:parents-selection}). Running SHAP from the trained models to select parents is done following a bootstrap approach, by sampling different portions of the data at each iteration to fit the selected Shapley explainer. After that, a filtered adjacency matrix (Section \ref{ss:adjacency-matrix-update}) is used to outline an unoriented graph, for each of the two regressors (Fig.~\ref{fig:summary}a,b). As a next step, the unoriented graphs become directed graphs (Fig.~\ref{fig:summary}c,d), by using ANM (HSIC) (Section \ref{ss:directing-edges}). These may contain edges that do not exist in the true DAG (e.g., $B \leftarrow C$ in (c)), or wrongly directed edges (e.g., $E \leftarrow C$ in (c) and $D \rightarrow C$ in (d)). The union of these two graphs results in the graph depicted in Fig. \ref{fig:summary}e, where all edges found by any one of the regressors are included. However, this union operation may result in some double-directed edges and cycles (e.g., $C \leftrightarrow D$, $C \leftrightarrow E$ and $B \rightarrow D \rightarrow C \rightarrow B$ in (e)) that need to be resolved (Section \ref{ss:final-dag}) to produce the final DAG (Fig. \ref{fig:summary}f).


\section{Results on synthetic data}\label{s:Results}

We first consider synthetic data for the evaluation, as it allows to operate under controlled assumptions, where the causal structure is fully specified, and the data generation process (see Appendix~\ref{app:synthetic-datasets}) is explicitly governed by known structural equations. Consequently, in this synthetic setup, there are no unobserved confounders (i.e., variables that are common causes of two or more measured variables which could induce spurious associations), thereby fulfilling the causal sufficiency assumption (which posits that all such common causes are indeed part of the observed dataset). The structural equations are also stable under interventions, while all other structural equations remain unchanged (i.e., actively setting a variable's value to observe its effects on downstream variables, as in \(P\bigl(Y \mid \mathrm{do}(X\!=\!x)\bigr)\)~\cite{pearl2009causality}). Furthermore, the data respects the faithfulness assumption (formally defined in Section~\ref{s:preliminaries} and Definition~\ref{def:faithfullness}), meaning all observed conditional independencies accurately reflect the d-separations in the true causal graph, with no accidental independencies due to parameter choices.

We conducted experiments using five families of synthetic datasets, each corresponding to a specific type of causal relationship: linear, polynomial, sigmoid additive, Gaussian additive, and Gaussian mixed models. These datasets vary in complexity and non-linearity, and were generated following the methodology described in \cite{kalainathanGenerativeNeuralNetworks2020} and \cite{Peters2014}.

For each synthetic generation mechanism family, we generated datasets with $p=10, 15, 20$ and $25$ variables, each containing $m=500$ samples (see Table~\ref{tab:median-stddev} for details on the distribution of number of edges in the ground truth DAGs). For a robust and reliable evaluation, 10 different datasets were generated for each combination of mechanism family and variable count, resulting in 200 datasets (5 families $\times$ 4 variable counts $\times$ 10 datasets). All synthetic datasets used to benchmark \methodname in this study have been made available in the GitHub repository, along the parametrization used for \methodname and compared methods.

These datasets, together with their corresponding unique ground truth DAGs, are used to benchmark \methodname against the state-of-the-art causal discovery methods (hereafter referred to as \textit{compared methods}) PC \cite{spirtes2000causation}, NOTEARS \cite{zheng2018dags}, GES \cite{chickering2002optimal}, LiNGAM \cite{shimizu2006linear}, FCI \cite{spirtes1999}, and CAM \cite{buhlmann2014cam}. NOTEARS was trained and thresholded with the optimum value found for all datasets used in the study. The rest of the methods have been used with default parametrization.

It is important to note that while these default configurations do not typically incorporate an explicit internal bootstrapping loop for edge stability in the manner of \methodname, our method's bootstrapping procedure (Algorithm~\ref{alg:dag-construction}) is an integral component designed to ensure the robustness of parent sets derived from SHAP values of its machine learning regressors (DFNs and GBTs), whose explanations can exhibit variability with data resampling.

All computations were performed on a system equipped with an Apple M2-Pro processor and 32 GB of RAM, without the use of parallelization or GPU acceleration.
\begin{table}
    \centering
    \caption{Average and standard deviation number of edges in the generated DAGs for $p=10, 15, 20, 25$.}
        \begin{tabular*}{0.77\textwidth}{lcccc}
            \toprule
            \textbf{Nr of features} & \textbf{10} & \textbf{15} & \textbf{20} & \textbf{25} \\
            \midrule
            Nr of edges in ground truth & $15 \pm 5.5$ & $23 \pm 7.62$ & $33 \pm 6.47$  & $42 \pm 10.31$ \\
            \bottomrule
        \end{tabular*}
    \label{tab:median-stddev}
\end{table}

\subsection{Evaluation metrics}

To compare a predicted Directed Acyclic Graphs (DAG) against a ground truth DAG, we first compute the True Positives (TPs), False Positives (FPs), and False Negatives (FNs). We define a TP as an edge with the correct direction in both the predicted and ground truth DAGs; a FP as an inferred edge that is not present, or has the opposite direction, in the ground truth DAG; and a FN as an edge present in the ground truth but missing or misoriented in the predicted DAG. Note that a true edge inferred but with the opposite direction is counted both as a FP and a FN. With these metrics, we then compute standard graph-based metrics: precision, recall, and F1-score. Precision is the ratio of correctly predicted edges (TPs) to all predicted edges (TPs + FPs), recall is the proportion of true edges recovered in the predicted DAG (i.e., TPs/(TPs+FNs)), and the F1-score is the harmonic mean of precision and recall. We also consider the difference in the number of edges between the predicted DAG and the true DAG, to evaluate whether a method tends to over- or under-estimate the number of edges. 

Additionally, we calculate the Structural Hamming Distance (SHD) \cite{tsamardinos2006max}, which measures the difference between two DAGs by counting missing or extraneous edges, and Structural Intervention Distance (SID) \cite{SID}, which evaluates how closely two DAGs agree on causal inference statements derived from interventions. Unlike SHD, SID focuses purely on the causal implications of the graph structure, making it particularly suitable for assessing causal discovery algorithms.

A high-performing DAG will demonstrate high precision, recall, and F1, alongside low SHD and SID values.

\subsection{Experimental results}

We first assess \methodname as a function of the number of variables $p$, across the five considered families of synthetic datasets, to evaluate its performance with increasingly complex datasets. 

\begin{figure*}[ht]
    \centering
    \begin{minipage}[t]{0.24\columnwidth}
        \centering
        \subfloat[]{\includegraphics[width=\linewidth]{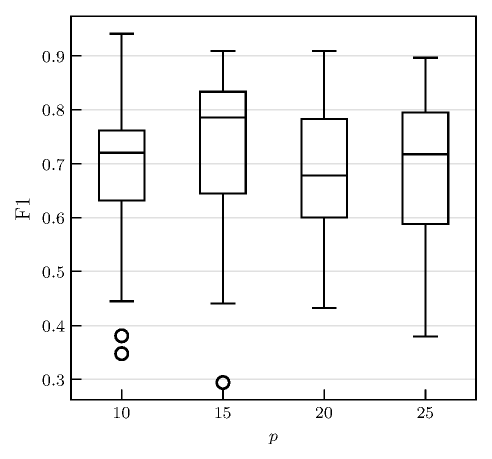}}
    \end{minipage}%
    \begin{minipage}[t]{0.24\columnwidth}
        \centering
        \subfloat[]{\includegraphics[width=\linewidth]{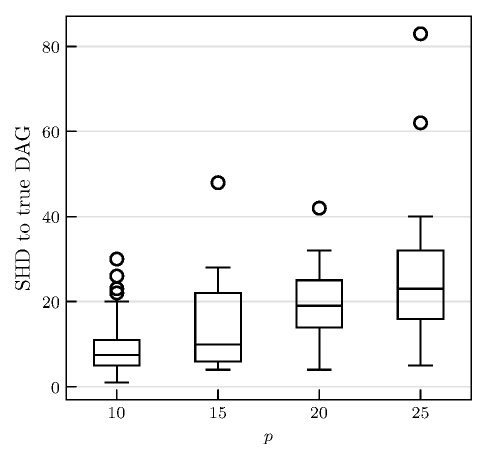}}
    \end{minipage}%
    \begin{minipage}[t]{0.24\columnwidth}
        \centering
        \subfloat[]{\includegraphics[width=\linewidth]{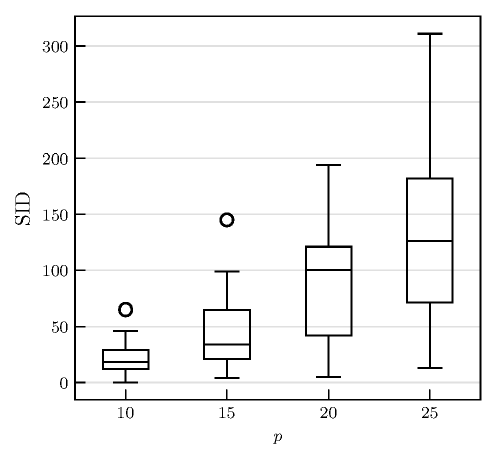}}
    \end{minipage}%
    \begin{minipage}[t]{0.24\columnwidth}
        \centering
        \subfloat[]{\includegraphics[width=\linewidth]{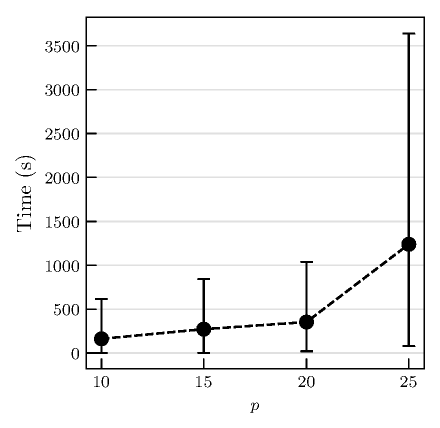}}
    \end{minipage}
    \caption{\footnotesize Effect of increasing the number of variables in the input dataset with \methodname in the following metrics: (a) F1 score; (b) SHD to true DAG; (c) SID to true DAG; and (d) computation time. Results are computed for the five considered families of synthetic data.
    }
    \label{fig:incresing-p}
\end{figure*}

Looking at the F1 score (Fig. \ref{fig:incresing-p}a), we observe that \methodname achieves high predictive performance, especially for lower and moderate values of \( p \), with median F1 scores consistently above 0.7. This reflects \methodname's capacity to balance precision and recall effectively in identifying causal relationships. Although some variability is present, the high F1 scores across different values of \( p \) underscore \methodname's robustness. However, there is a slight decrease in the median F1 score for the highest value of \( p \), suggesting that scalability might pose a limitation to its predictive power in very high-dimensional settings. 

In terms of SHD (Fig. \ref{fig:incresing-p}b), \methodname exhibits relatively low SHD values, with a concentrated distribution, particularly for smaller values of \( p \). This suggests that \methodname maintains structural accuracy effectively, even as the dimensionality grows. However, there is a noticeable upward trend in SHD for larger values of \( p \), indicating that while \methodname performs well in smaller graphs, its structural precision may gradually decrease as the dimensionality increases. As for SID (Fig. \ref{fig:incresing-p}c), we also see remarkable low values for lower $p$ and a similar trend to that shown for SHD, as the number of variables increases.

Next, we evaluate the computation time (Fig. \ref{fig:incresing-p}d), showing that \methodname’s time requirements increase as \( p \) grows, which is an expected outcome given the combinatorial nature of SHAP values computation. While the computational cost becomes more significant at higher values of \( p \), the performance remains reasonable for moderate-dimensional datasets. Users should be aware that runtime will increase substantially with higher-dimensional data, indicating a potential challenge in scalability.

Overall, \methodname demonstrates strong performance in terms of structural accuracy and F1 score, especially for small to medium-sized datasets, maintaining computational demands and structural precision acceptable for moderate values of \( p \).

\subsubsection{Results by synthetic dataset}

When evaluating these metrics per synthetic dataset family (see Appendix \ref{appendix:add-results}, Fig.~\ref{fig:additional-metrics}), we do not observe significant differences. The results indicate that \methodname performs robustly across a range of complexities in data structure, showing consistent trends across various settings.

\methodname achieves high F1 scores (Fig.~\ref{fig:additional-metrics}a) across all dataset families, with particularly strong performance in the Gaussian (additive), Gaussian (mixed), and Sigmoid (additive) families, where F1 scores remain above 0.6 even as \( p \) increases. In contrast, the Linear and Polynomial datasets show more variability, particularly at higher \( p \) values, indicating greater difficulty for \methodname in these simpler structures. Precision and recall (Fig.~\ref{fig:additional-metrics}b,c) follow similar trends, with consistently high values in the Gaussian and Sigmoid families, reflecting \methodname’s ability to capture relevant causal relationships effectively. Linear and Polynomial datasets, however, show greater variability, suggesting that \methodname may struggle with these structured relationships at higher dimensionalities.

The SHD and SID metrics (Fig.~\ref{fig:additional-metrics}d,e) reinforce these findings, with low values in Gaussian and Sigmoid families, indicating \methodname’s accuracy in reconstructing causal structures and capturing correct causal directions. Higher values in Linear and Polynomial datasets, particularly as \( p \) increases, suggest occasional challenges in directionality and structure for these settings.

Training time (Fig.~\ref{fig:additional-metrics}f) increases with \( p \) across all families, as expected. \methodname requires less time on Linear and Polynomial datasets compared to Gaussian and Sigmoid, likely due to the simpler relationships in these structures.

\subsubsection{Comparison with other methods}

\begin{figure*}[ht]
    \centering
    \begin{minipage}[t]{0.4\columnwidth}
        \centering
        \subfloat[]{\includegraphics[width=\linewidth]{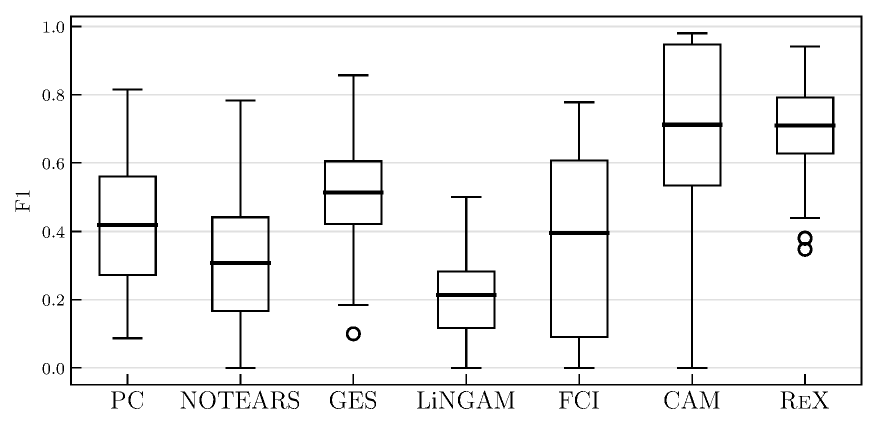}}
    \end{minipage}%
    \begin{minipage}[t]{0.4\columnwidth}
        \centering
        \subfloat[]{\includegraphics[width=\linewidth]{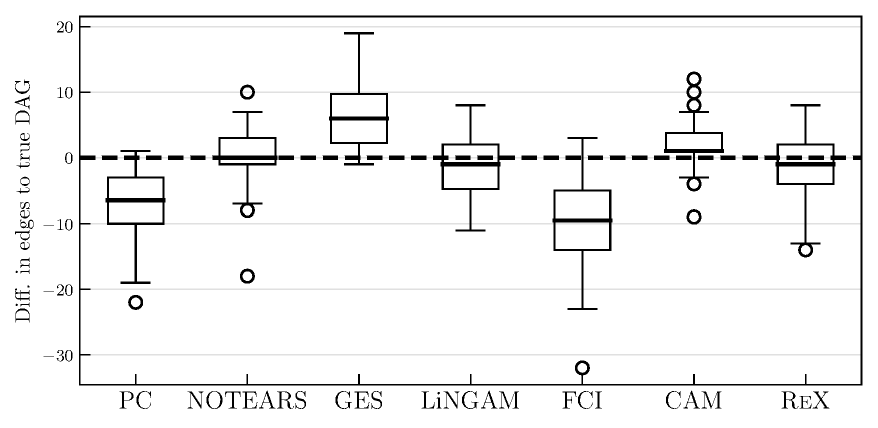}}
    \end{minipage}
    \begin{minipage}[t]{0.4\columnwidth}
        \centering
        \subfloat[]{\includegraphics[width=\linewidth]{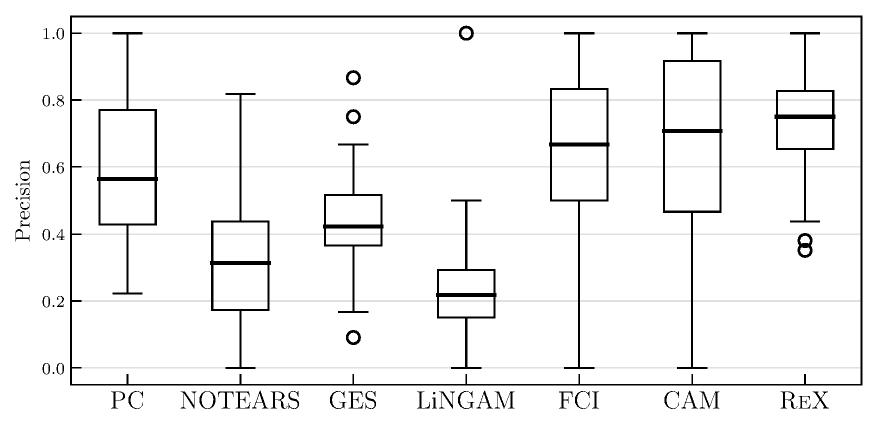}}
    \end{minipage}%
    \begin{minipage}[t]{0.4\columnwidth}
        \centering
        \subfloat[]{\includegraphics[width=\linewidth]{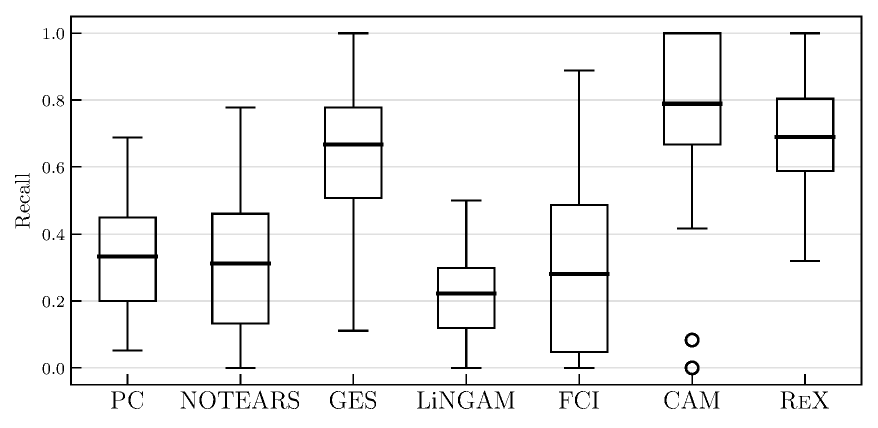}}
    \end{minipage}
    \caption{\footnotesize Comparative performance of \methodname against the benchmark methods PC, NOTEARS, GES, LiNGAM, FCI, and CAM across four key metrics: (a) F1 score; (b) edge difference, (c) precision; and (d) recall. Metrics are computed across all five considered synthetic families, with $p = 10$ features.}
    \label{fig:metrics-compared}
\end{figure*}

\begin{figure*}[ht]
    \centering
    \begin{minipage}[t]{0.4\columnwidth}
        \centering
        \subfloat[]{\includegraphics[width=\linewidth]{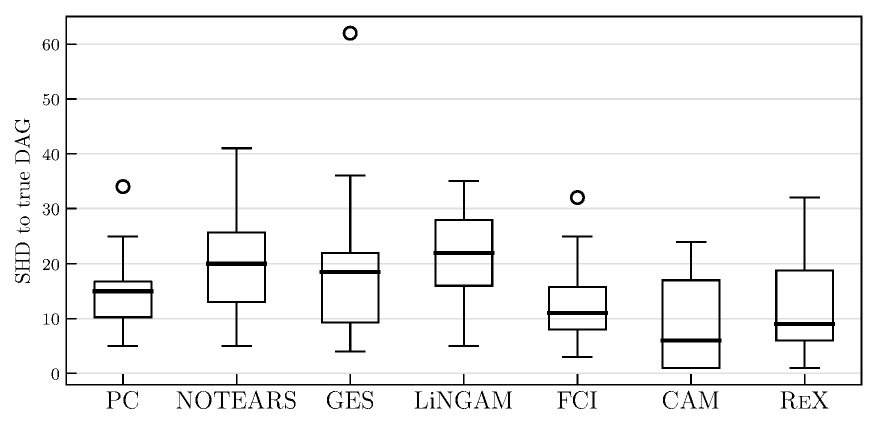}}
    \end{minipage}%
    \begin{minipage}[t]{0.4\columnwidth}
        \centering
        \subfloat[]{\includegraphics[width=\linewidth]{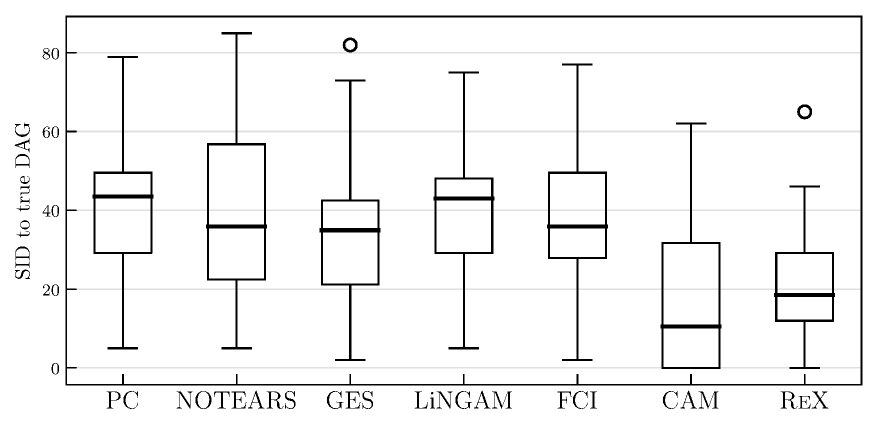}}
    \end{minipage}
    \caption{\footnotesize Benchmark between \methodname and compared methods for (a) SHD to true DAG and (b) SID to true DAG metrics.}
    \label{fig:shd-sid}
\end{figure*}

Next, we compare \methodname against the compared methods (see also Appendix \ref{app:sample-dags} for sample DAGs generated by each method). We observe that \methodname consistently shows strong performance, particularly in F1 score (Fig.~\ref{fig:metrics-compared}a), where it achieves higher values relative to most other methods. Notably, \methodname exhibits a lower variance than the other methods, with values tightly concentrated around the median. This stability across metrics suggests that \methodname is less sensitive to variations in the data, resulting in more consistent performance. As suggested by the obtained F1 scores, \methodname stands out in aligning well with the true number of edges in the causal graph (Fig. \ref{fig:metrics-compared}b), indicating its effectiveness at avoiding both excessive edge predictions (false positives) and missing edges (false negatives). Compared to other methods, it exhibits a narrower interquartile range, implying lower variability in edge prediction across different datasets. This consistency suggests a robustness in \methodname’s performance, with fewer instances of significant deviations from the true edge count. The wider spread in edge difference for some of the compared methods observed in Figures~\ref{fig:metrics-compared} and \ref{fig:shd-sid} reflect their operation under default parameterizations (as stated in Section~\ref{s:Results}). These default settings might not be universally optimized for sparsity across all diverse synthetic data generation processes we tested, potentially leading them to predict denser graphs in certain scenarios compared to \methodname's intrinsic mechanisms for controlling graph density.

In terms of precision (Fig.~\ref{fig:metrics-compared}c), \methodname performs comparably well with methods like CAM and FCI, demonstrating that it avoids excessive false positives while capturing meaningful causal links. This level of precision is advantageous for applications requiring high-confidence causal relationships. However, \methodname’s performance in recall (Fig.~\ref{fig:metrics-compared}d) is moderate, aligning closely with other methods like PC and GES, indicating that while \methodname is effective in identifying relevant causal relationships, it may miss some causal connections compared to approaches with higher recall like CAM.

Overall, \methodname demonstrates strong structural accuracy and a balanced detection capability, particularly excelling in structural precision and F1 score, which are indicative of its robustness in maintaining the validity of discovered causal structures. However, some trade-offs in recall suggest areas where further improvements could be explored to enhance \methodname's sensitivity in capturing all causal connections, especially in more complex causal graphs.

In addition, \methodname achieves competitive SHD values (Fig.~\ref{fig:shd-sid}a), with a median lower than most compared methods, indicating good alignment with the true DAG structure. The variability in SHD for \methodname is moderate, showing some datasets where performance is comparable to CAM and FCI, while outperforming methods like LiNGAM and GES. These results suggest that \methodname effectively balances its structural recovery, minimizing the number of missing and extra edges across diverse scenarios.

In terms of SID, \methodname shows favorable performance (Fig.~\ref{fig:shd-sid}b), with lower values than most methods and a narrower interquartile range compared to GES and NOTEARS. This indicates that \methodname better captures interventionally relevant causal directions, providing robust predictions under interventions. While CAM achieves comparable SID performance, \methodname demonstrates a more consistent behavior across datasets, highlighting its adaptability to various data structures. These results confirm \methodname's effectiveness in constructing causal graphs that align closely with both the observational and interventional ground truth.

\section{Evaluation on biological and financial complex systems} \label{s:results-complex-systems}

\subsection{Single-cell protein network data}

To evaluate \methodname's performance on a real-world biological system and facilitate comparison within the broader causal discovery literature, we selected the Sachs et al. (2005) single-cell protein-signaling network dataset \cite{sachs2005causal}. This dataset is a prominent benchmark due to several factors: it represents a complex, genuine biological system; it possesses a widely accepted, albeit simplified, ground truth causal network derived from experimental interventions, which is crucial for quantitative validation; and its inherent characteristics, such as potential non-linear interactions, make it a relevant testbed for methods like \methodname designed to handle such complexities. For our analysis, we utilize the cleaned version of the Sachs dataset, as provided in \cite{ramsey2018fask}, and used in \cite{chang2024postselectioninferencecausaleffects}.

Following the structure described in Sachs et al. (2005) and Ramsey and Andrews (2018), we divide the 11 variables into two tiers: Plc, Pkc, Pka, PIP2, and PIP3 are categorized into tier 1, while the remaining variables (Raf, Mek, Erk, Akt, P38, and Jnk) belong to tier 2. Grouping variables in tiers is done to represent a causal dependency (temporal) order, such that variables at a given tier (e.g., $1$) are not expected to receive directed edges from those in subsequent tiers (i.e., $2, 3, \ldots$), but only from variables in the same or previous tiers. Hence, although feedback loops are suggested to exist within the network, they are thought to primarily occur among variables in tier 1, with no directed edges expected from tier 2 to tier 1. This prior information is added to \methodname and the compared methods prior to running them.

\begin{figure}[ht!]
    \centering
    \includegraphics[width=0.32\textwidth]{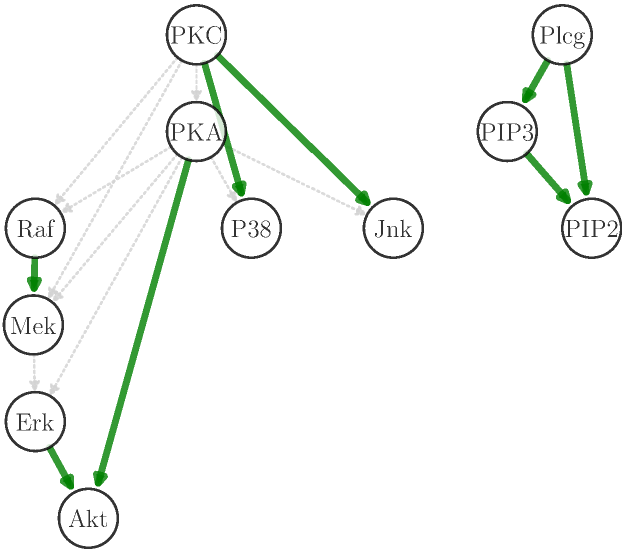}
    \caption{Plausible causal graph obtained by \methodname on the Sachs dataset. Green arrows correspond to correctly predicted edges, while gray arrows are true edges missed by \methodname.}
    \label{fig:sachs-long}
\end{figure}

Fig. \ref{fig:sachs-long} shows the causal graph obtained by \methodname on the Sachs dataset. With a precision of $0.952$, a recall of $0.471$ and a F1-score of $0.629$, \methodname is able to correctly recover important causal relationships from the data. The number of wrong edges or wrong directions are minimized, though the entire set of edges is not recovered. We note that to recover this DAG, \methodname was applied with exactly the same parametrization as in the synthetic datasets.

\begin{table}
    \caption{Summary metrics for all considered methods on the Sachs dataset.}
    \centering
    \begin{tabular*}{0.64\textwidth}{@{} lccccc @{}}
        \toprule
        \textbf{Method} & \textbf{Precision} & \textbf{Recall} & \textbf{F1} & \textbf{SHD} & \textbf{SID} \\
        \midrule
        PC          & $0.462 \scalemath{0.6}{\ \pm 0.04}$ & $0.316 \scalemath{0.6}{\ \pm 0.05}$ & $0.376 \scalemath{0.6}{\ \pm 0.04}$  & $15 \scalemath{0.6}{\ \pm 2}$ &   $53 \scalemath{0.6}{\ \pm 3}$ \\
        FCI         & $0.667 \scalemath{0.6}{\ \pm 0.05}$ & $0.421 \scalemath{0.6}{\ \pm 0.06}$ & $0.516 \scalemath{0.6}{\ \pm 0.05}$  & $13 \scalemath{0.6}{\ \pm 3}$ &   $\bm{36} \scalemath{0.6}{\ \pm 4}$ \\
        GES         & $0.200 \scalemath{0.6}{\ \pm 0.02}$ & $0.421 \scalemath{0.6}{\ \pm 0.03}$ & $0.273 \scalemath{0.6}{\ \pm 0.03}$  & $41 \scalemath{0.6}{\ \pm 1}$ &   $50 \scalemath{0.6}{\ \pm 2}$ \\
        LiNGAM      & $0.115 \scalemath{0.6}{\ \pm 0.01}$ & $0.158 \scalemath{0.6}{\ \pm 0.02}$ & $0.134 \scalemath{0.6}{\ \pm 0.02}$  & $37 \scalemath{0.6}{\ \pm 1}$ &   $53 \scalemath{0.6}{\ \pm 1}$ \\
        NOTEARS     & $0.212 \scalemath{0.6}{\ \pm 0.03}$ & $0.368 \scalemath{0.6}{\ \pm 0.04}$ & $0.270 \scalemath{0.6}{\ \pm 0.03}$  & $36 \scalemath{0.6}{\ \pm 2}$ &   $50 \scalemath{0.6}{\ \pm 2}$ \\
        CAM         & $0.250 \scalemath{0.6}{\ \pm 0.05}$ & $\bm{0.600} \scalemath{0.6}{\ \pm 0.06}$ & $0.353 \scalemath{0.6}{\ \pm 0.05}$  & $39 \scalemath{0.6}{\ \pm 3}$ &   $47 \scalemath{0.6}{\ \pm 4}$ \\
        \textbf{\methodname} & $\bm{0.952} \scalemath{0.6}{\ \pm 0.02}$ & $0.471 \scalemath{0.6}{\ \pm 0.01}$ & $\bm{0.629} \scalemath{0.6}{\ \pm 0.02}$ & $\bm{9} \scalemath{0.6}{\ \pm 1}$ & $39 \scalemath{0.6}{\ \pm 2}$ \\
        \bottomrule
    \end{tabular*}
    \label{tab:sachs-metrics}
\end{table}

We compare \methodname's performance with that of the compared methods on the Sachs dataset (Table~\ref{tab:sachs-metrics}). The results are based on five-runs with different seeds, all methods using parameter settings consistent with those applied to the synthetic datasets.

\begin{figure}[htp!]
    \centering
    \begin{minipage}{0.8\textwidth}
    \centering
    \begin{minipage}{0.32\textwidth}
        \centering
        \includegraphics[width=\linewidth]{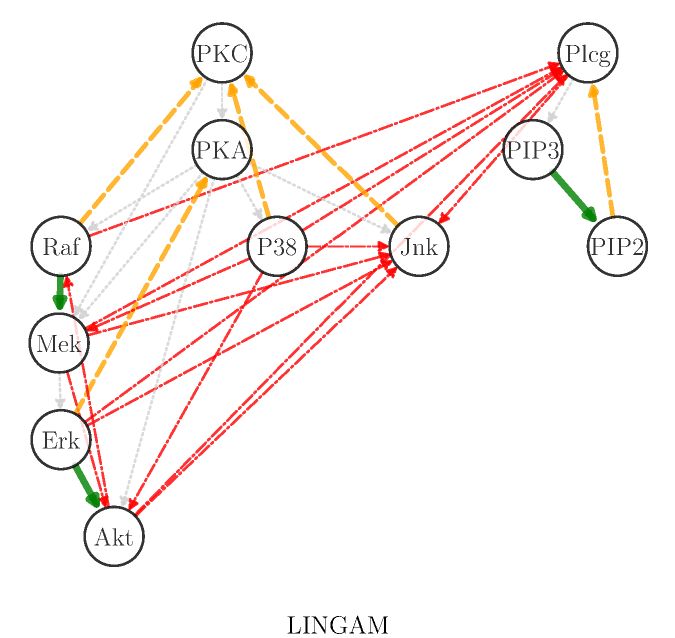}
    \end{minipage}\hfill 
    \begin{minipage}{0.32\textwidth}
        \centering
        \includegraphics[width=\linewidth]{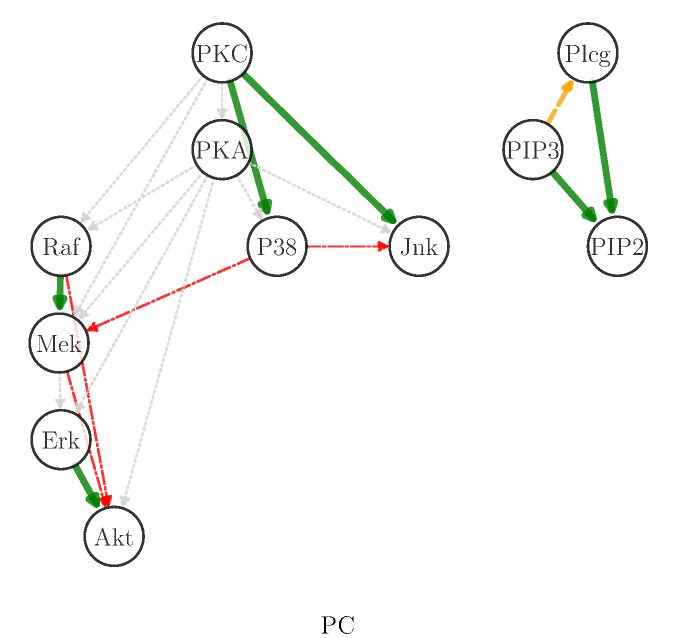}
    \end{minipage}\hfill
    \begin{minipage}{0.32\textwidth}
        \centering
        \includegraphics[width=\linewidth]{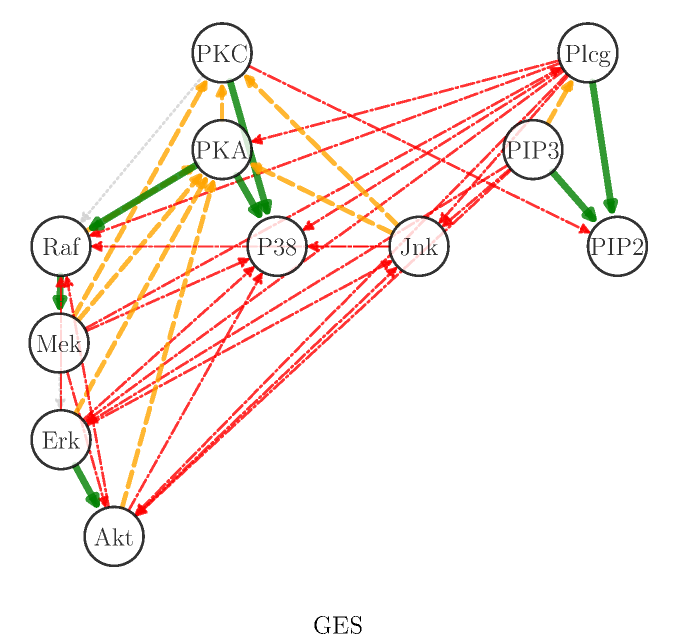}
    \end{minipage}

    \vspace{0.5cm} 

    \begin{minipage}{0.32\textwidth}
        \centering
        \includegraphics[width=\linewidth]{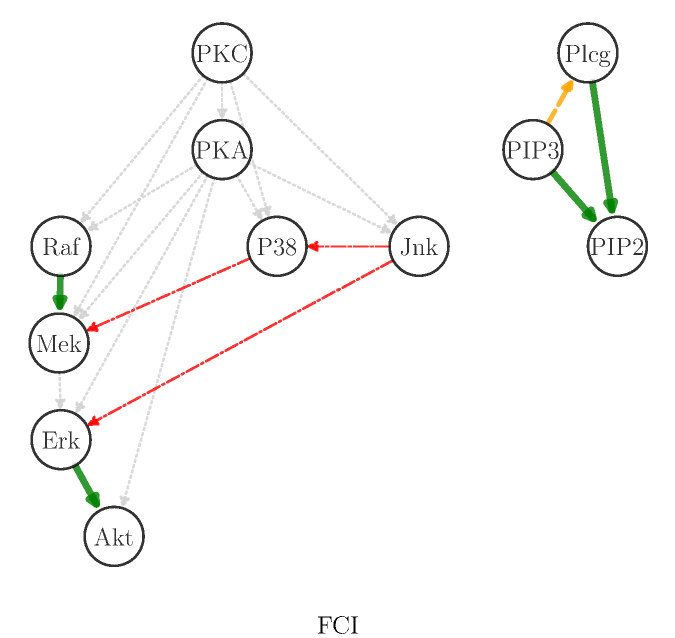}
    \end{minipage}\hfill
    \begin{minipage}{0.32\textwidth}
        \centering
        \includegraphics[width=\linewidth]{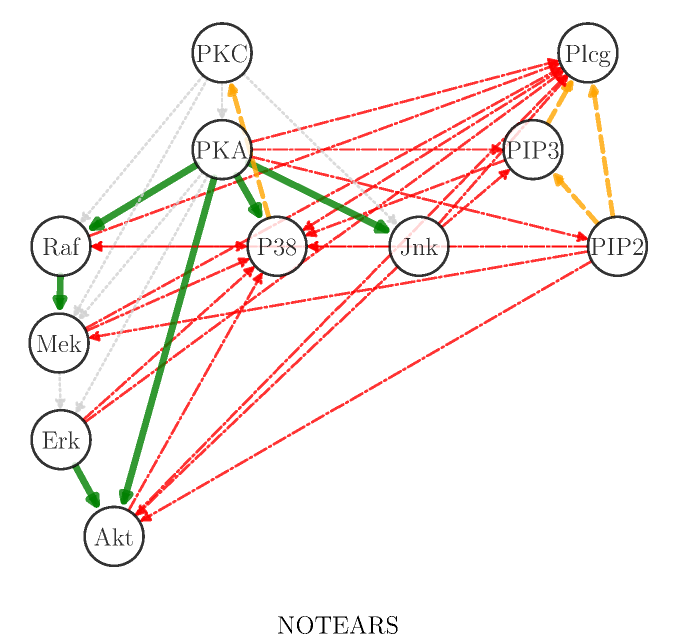}
    \end{minipage}\hfill
    \begin{minipage}{0.32\textwidth}
        \centering
        \includegraphics[width=\linewidth]{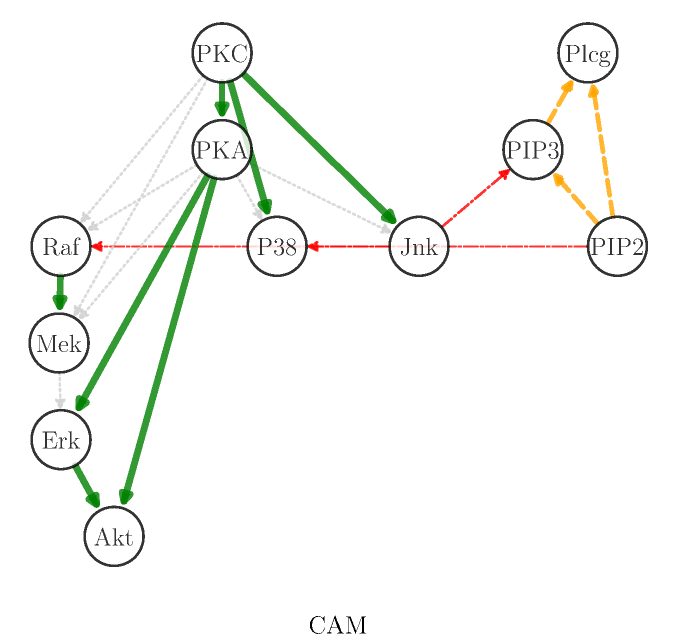}
    \end{minipage}
    \caption{DAGs obtained by the rest of comparable methods on the Sachs datasets.}
    \label{fig:sachs-other-methods} 
    \end{minipage}
\end{figure}

The results in Table~\ref{tab:sachs-metrics} demonstrate that \methodname significantly outperforms other methods across most evaluation metrics on this dataset, achieving the highest score in precision and F1 score, indicating both high specificity and a strong balance between precision and recall. While its recall of $0.471$ is moderate, it is still competitive and notably higher than that of most methods, except for CAM, which achieved a higher recall of $0.6$. In terms of structural accuracy, \methodname also reports the lowest SHD (median of $9$), implying fewer structural errors compared to others, with FCI and CAM following at $13$ and $39$, respectively. In addition, \methodname has a relatively good value for SID (median of $39$), indicating also a noteworthy intervention-based structure interpretation capability. Overall, \methodname demonstrates strong precision and structural fidelity, supporting its robustness and reliability for causal discovery in this context (see Figure~\ref{fig:sachs-other-methods}).

While it is possible that further tuning could improve the performance of other methods, \methodname achieves top scores in precision and F1 without such adjustments, highlighting its robustness and reliability in causal discovery under default settings.

\subsection{Financial decision-making data} \label{s:results-financial-data}

\begin{figure*}[ht]
    \centering
    \begin{minipage}[t]{0.32\columnwidth}
        \centering
        \subfloat[]{\includegraphics[width=\linewidth]{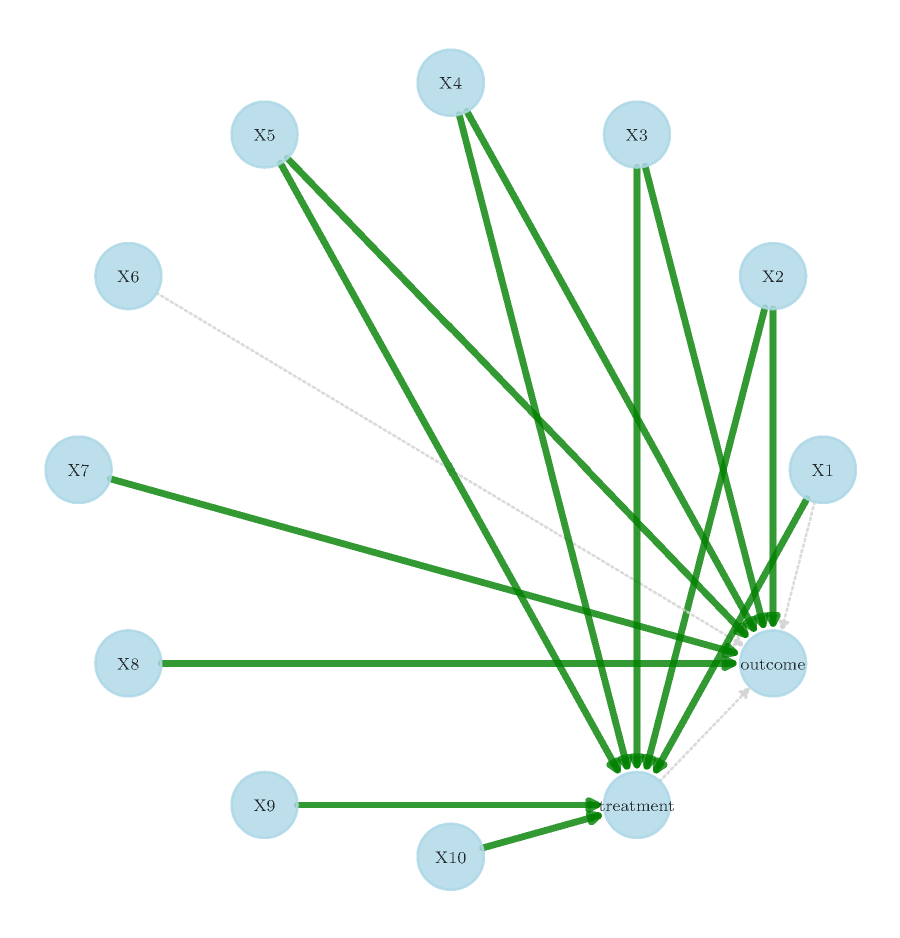}}
    \end{minipage}%
    \begin{minipage}[t]{0.32\columnwidth}
        \centering
        \subfloat[]{\includegraphics[width=\linewidth]{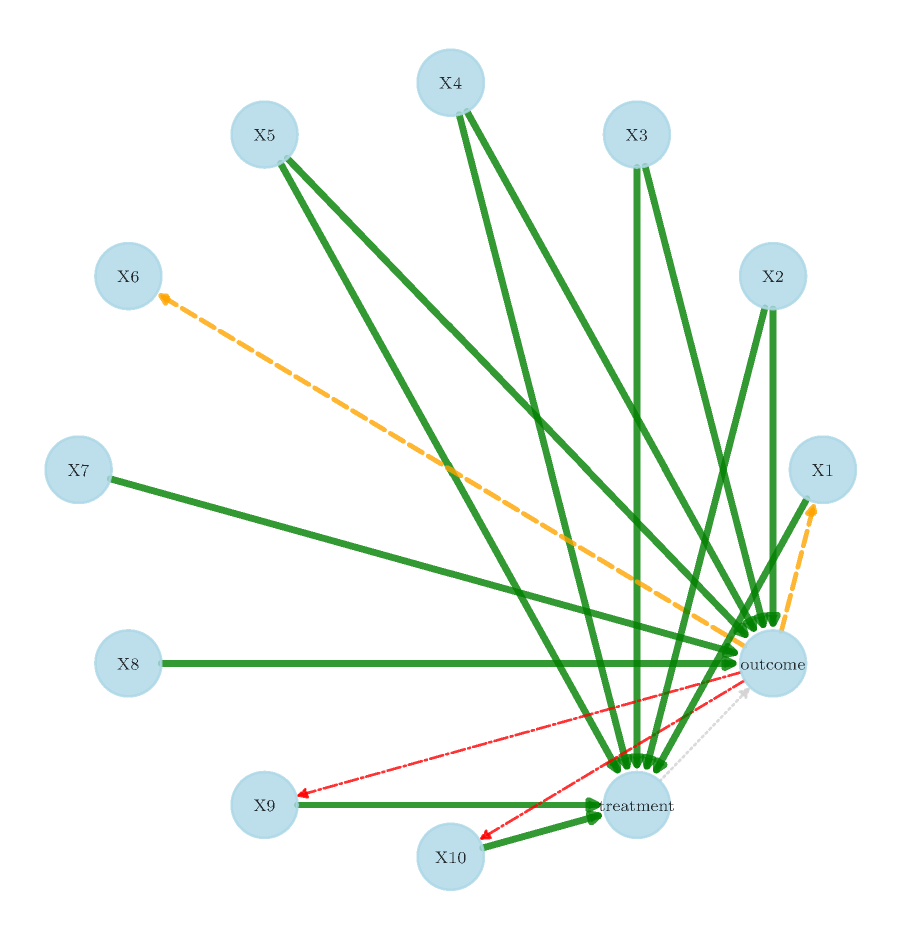}}
    \end{minipage}
    \begin{minipage}[t]{0.32\columnwidth}
        \centering
        \subfloat[]{\includegraphics[width=\linewidth]{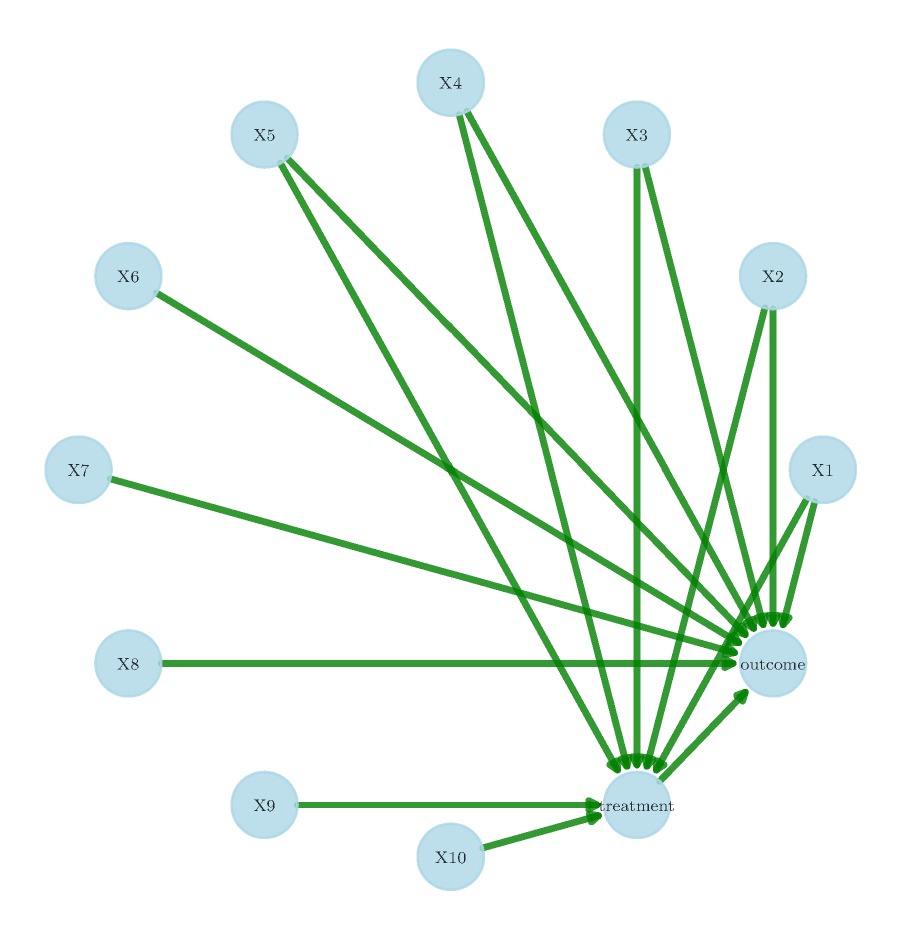}}
    \end{minipage}
    \caption{\footnotesize Two samples of the DAGs inferred by \methodname on the financial dataset: (a) the DAG with the best inferred structure, and (b) the DAG with the worst inferred structure, as well as (c) the true DAG.}
    \label{fig:finance-sample-dags}
\end{figure*}

The \methodname method was further evaluated on a synthetic financial dataset \cite{causal_debt_datagen} designed to emulate real-world financial decision-making scenarios. This dataset captures the structural complexities typically encountered in financial credit risk assessments, including heterogeneous treatment effects, non-linear interactions, and positivity assumption violations. The dataset consists of 12 variables, with an observational sample of 12,000 entries. The variables include financial indicators such as credit history, debt levels, number of loans, and external debt exposure. The treatment variable represents the percentage of debt write-off assigned to customers, while the outcome variable reflects repayment probability. The known true causal DAG (see Fig. \ref{fig:finance-sample-dags}c) underlying the data generation process serves as ground truth for evaluating the effectiveness of different causal discovery methods (see \cite{causal_debt_datagen} for details on the data generation process).

The causal discovery methods were applied ten times to different data subset samples (made of 1,000 samples), and performance was assessed based on precision, recall, F1-score, SHD and SID. \methodname was applied following the same pipeline as in previous experiments. As with the single-cell protein network data, a hyperparameter search (300 iterations) was conducted to optimize the regressors used for SHAP value computation. Each dataset was split into training and validation subsets, with the causal discovery process applied to the entire dataset after model training. Bootstrapping (10 iterations) was employed to enhance stability in the inferred causal structure, for the \methodname case. Two sample DAGs (best and worst) extracted from the bootstrapping process are show in Figure~\ref{fig:finance-sample-dags}, in addition to the ground truth.
\begin{table}[ht]
    \caption{Summary metrics for all considered methods on the Financial dataset.}
    \centering
    \begin{tabular*}{0.7\textwidth}{@{} lccccc @{}}
        \toprule
        \textbf{Method} & \textbf{Precision} & \textbf{Recall} & \textbf{F1} & \textbf{SHD} & \textbf{SID} \\
        \midrule
        PC      & $0.950   \scalemath{0.6}{\ \pm 0.11}$ & $0.375 \scalemath{0.6}{\ \pm 0.10}$ & $0.530 \scalemath{0.6}{\ \pm 0.11}$ & $16.0 \scalemath{0.6}{\ \pm 0.0}$ & $13.0 \scalemath{0.6}{\ \pm 1.58}$ \\
        FCI     & $\bm{0.967}  \scalemath{0.6}{\ \pm 0.07}$ & $0.375 \scalemath{0.6}{\ \pm 0.12}$ & $0.533 \scalemath{0.6}{\ \pm 0.12}$ & $10.2 \scalemath{0.6}{\ \pm 2.0}$ & $13.0 \scalemath{0.6}{\ \pm 1.87}$ \\
        GES     & $0.659  \scalemath{0.6}{\ \pm 0.09}$ & $0.575 \scalemath{0.6}{\ \pm 0.07}$ & $0.610 \scalemath{0.6}{\ \pm 0.06}$ & $11.8 \scalemath{0.6}{\ \pm 2.5}$ & $34.6 \scalemath{0.6}{\ \pm 9.61}$ \\
        LiNGAM  & $0.227  \scalemath{0.6}{\ \pm 0.11}$ & $0.137 \scalemath{0.6}{\ \pm 0.08}$ & $0.170 \scalemath{0.6}{\ \pm 0.09}$ & $21.0 \scalemath{0.6}{\ \pm 2.5}$ & $19.0 \scalemath{0.6}{\ \pm 0.00}$ \\
        NOTEARS & $0.558  \scalemath{0.6}{\ \pm 0.10}$ & $0.562 \scalemath{0.6}{\ \pm 0.06}$ & $0.553 \scalemath{0.6}{\ \pm 0.03}$ & $14.6 \scalemath{0.6}{\ \pm 2.2}$ & $9.2  \scalemath{0.6}{\ \pm 1.30}$ \\
        CAM     & $0.145  \scalemath{0.6}{\ \pm 0.11}$ & $0.212 \scalemath{0.6}{\ \pm 0.12}$ & $0.171 \scalemath{0.6}{\ \pm 0.11}$ & $35.4 \scalemath{0.6}{\ \pm 9.1}$ & $83.4 \scalemath{0.6}{\ \pm 15.04}$ \\
        \textbf{\methodname} & $0.915 \scalemath{0.6}{\ \pm 0.12}$ & $\bm{0.799} \scalemath{0.6}{\ \pm 0.03}$ & $\bm{0.850} \scalemath{0.6}{\ \pm 0.05}$ & $\bm{4.6} \scalemath{0.6}{\ \pm 1.82}$ & $\bm{6.4} \scalemath{0.6}{\ \pm 1.52}$ \\
        \bottomrule
    \end{tabular*}
    \label{tab:finance-metrics}
\end{table}

The summary metrics for all considered methods are provided in Table~\ref{tab:finance-metrics}. The results indicate that \methodname consistently outperforms other methods in terms of structural accuracy. With an F1-score of $0.850 \pm 0.05$, \methodname surpasses the second-best performing method, GES, which achieves an F1-score of $0.610 \pm 0.06$. This improvement can be attributed to \methodname’s ability to effectively leverage machine learning explainability techniques for causal inference.

In terms of precision, \methodname achieves a score of $0.915 \pm 0.12$, which is only slightly lower than the highest-performing method in this regard, FCI, which reaches $0.967 \pm 0.07$. However, \methodname excels in recall with a significantly higher score of $0.799 \pm 0.03$, substantially outperforming FCI and PC, both of which yield recall values of $0.375 \pm 0.12$ and $0.375 \pm 0.10$, respectively. This suggests that \methodname is more effective at identifying true causal relationships without omitting key connections.

Furthermore, \methodname achieves the lowest Structural Hamming Distance (SHD) and Structural Intervention Distance (SID) among all tested methods, with values of $4.6 \pm 1.82$ and $6.4 \pm 1.52$, respectively. These results highlight the method’s ability to recover the true causal structure with minimal errors and to maintain robustness in intervention-based causal queries.

Compared to constraint-based methods such as PC and FCI, \methodname shows superior performance in terms of both recall and SHD. The performance of NOTEARS and LiNGAM remains significantly lower, with F1-scores of $0.553 \pm 0.03$ and $0.170 \pm 0.09$, respectively. The CAM method performs the worst, with an F1-score of $0.171 \pm 0.11$, an SHD of $35.4 \pm 9.1$, and the highest SID value of $83.4 \pm 15.04$, indicating that it struggles to correctly infer causal relationships in this domain.

\section{Practical considerations and future work} \label{s:discussion}

During the development of \methodname, we addressed several considerations and decisions, while also identifying potential areas for future improvement. Next, we discuss in detail topics such as opting for SHAP values over other explainability methods, the challenges of multicollinearity, and the potential for experimenting with various machine learning regressors. These factors play a crucial role in shaping the method's performance and open up opportunities for further research to refine and enhance the proposed approach.

\subsection{Selection of Shapley values for feature importance assessment}
\label{ss:choice-of-shapley}

A fundamental design decision in \methodname involves selecting a reliable measure of feature contributions to predict a target variable, which directly guides parent selection (Algorithm~\ref{alg:parent-selection}). Among various methods, Shapley values \cite{shapley1953value} were chosen due to their theoretical rigor and practical advantages in identifying stable and meaningful candidate parents.

Alternative methods have specific limitations. Model-based metrics, such as Gini importance, may bias results toward high-cardinality features or produce misleading attributions when features are correlated \cite{Strobl2007BiasRF}. Permutation importance, although intuitive, struggles similarly in correlated scenarios, either underestimating collective feature importance or producing unrealistic permutations \cite{Hooker2019UnrestrictedPI, Molnar2022Book}. Local explanation methods like LIME \cite{Ribeiro2016LIME} require extensive aggregation for global insights and are sensitive to the choice of local neighborhoods and surrogate models \cite{AlvarezMelis2018LIMEInstability}.

Conversely, Shapley values as implemented through SHAP \cite{lundberg2017unified} provide several distinct benefits for \methodname: (i) they uniquely satisfy key axiomatic properties (efficiency, symmetry, and additivity) that enable principled attribution \cite{shapley1953value, Winter2002Shapley}; (ii) they inherently accommodate feature interactions and redundancies by averaging contributions across all possible feature coalitions, ensuring stable attribution even in scenarios of high multicollinearity, as demonstrated empirically (Appendix~\ref{app:shapley-validation}); (iii) they exhibit a theoretical and empirically validated link to conditional dependence (Section~\ref{ss:mathematical-foundation}, Appendix~\ref{app:shapley-validation}), aligning closely with the requirements of causal discovery under faithfulness; and (iv) they benefit from robust, model-specific implementations (e.g., TreeExplainer for GBTs, GradientExplainer for DFNs), improving computational efficiency and reliability \cite{lundberg2017unified, chen_true_2020}.

Due to their theoretical soundness, robustness to feature correlations, and clear interpretability regarding conditional dependence, Shapley values are particularly well-suited to identifying potential causal parents in \methodname.

\subsection{Multicollinearity considerations} \label{ss:Multicollinearity}

A significant challenge when using Shapley values for model interpretation, particularly in the context of establishing causal relationships among features in a dataset, comes from situations where some of these features are highly correlated \cite{janzing_feature_2020}. An scenario with multicollinearity makes difficult to isolate the individual effect of each feature on the model's predictions. In such cases, Shapley values may be argued to distribute the importance across correlated features, potentially obscuring the true contribution of each individual feature to the model's predictions. This distribution of importance can be understood as source of ambiguity, as it could become challenging to discern whether the impact on the prediction is due to the intrinsic value of a feature or its correlation with other features.

The implementation of Shapley values in SHAP by \cite{lundberg2017unified} includes model-agnostic explainers like GradientExplainer or TreeExplainer as a potential solution \cite{chen_true_2020} to mitigate the effects of multicollinearity.

Specifically, these ``interventional'' variants of SHAP explainers, as discussed by Chen et al. \cite{chen_true_2020} in contrast to ``observational'' approaches that use conditional expectations, aim to estimate a feature's effect by breaking its correlations with other features not currently in the evaluated coalition. When computing the expected model output for a subset of features $S$, features $X_k \notin S$ are effectively marginalized out by sampling from their empirical distribution (often a background dataset) independently of the features in $S$, rather than conditioning on them.

For instance, \textbf{TreeExplainer}, when applied to tree-based models like GBTs, can implement this through path-dependent feature perturbation algorithms that effectively average over the marginal distribution of ``absent'' features encountered along decision paths. \textbf{GradientExplainer}, used for DFNs, approximates SHAP values and typically relies on a background dataset to represent features not in the current coalition; if these are sampled independently, its behavior aligns with an interventional perspective. 

Unlike other explainers that might treat features collectively or consider their joint distributions, interventional explainers operate under an interventional framework where the impact of each feature on the model's output is computed independently, assuming other features remain constant. This approach aligns with the concept of partial derivatives, focusing on the sensitivity of the output to changes in each feature, treated in isolation. By doing so, interventional explainers effectively disentangle the intertwined influences of correlated features, making it easier to interpret the model in terms of causal relationships. This is particularly useful in deep learning models where the intricate interactions of features can be complex and opaque.

In the context of \methodname, the adoption of model-agnostic explainers is intricately linked to the type of regressor being employed. When utilizing DFN, GradientExplainer is chosen, tailored to capture the complex, layered interactions and non-linear mappings characteristic of these models. Conversely, when employing Gradient Boosting Trees, TreeExplainer is utilized as it is more suited to dissecting the sequential, decision-tree-based learning process intrinsic to these algorithms. This adaptability in the choice of explainers is crucial, as it ensures that the explanations generated are not only robust but also appropriately aligned with the underlying mechanics of the chosen regression model.

Indeed, our synthetic experiments (detailed in Appendix \ref{app:shapley-validation}) provide empirical evidence for this robustness. In scenarios with highly collinear predictor variables, the Shapley values computed remained stable across repeated repeated trials ($\sigma/\mu < 5 \%$ for $\phi$ across 50 replicates) and provided an intuitive partitioning of importance between the correlated features, a task that can be challenging for other interpretation methods or even for establishing stable DAG structures using iterative CI-tests.

\subsection{Stability of the discovered DAG} \label{ss:stability-dag} 

The stability of the causal graph discovered by \methodname\ is primarily addressed through the bootstrapping mechanism (Algorithm~\ref{alg:dag-construction}) and the subsequent frequency-based thresholding using the $\tau$ parameter (Section~\ref{ss:adjacency-matrix-update}). An edge is incorporated into the initial graph for each regressor type ($G_{DFN}$ or $G_{GBT}$) only if its selection frequency across $T$ bootstrap iterations surpasses $\tau$. This procedure ensures that the foundational structure is based on relationships that consistently emerge despite resampling of the data, thereby filtering out less robust, potentially spurious connections.

While \methodname\ does not compute a separate, global quantitative metric for DAG stability over multiple identical full runs, the design inherently promotes the stability of the identified edges. The consistency and relatively low variance observed in performance metrics across diverse synthetic datasets (e.g., F1 scores in Figure~\ref{fig:additional-metrics}) further provide indirect evidence of the method's ability to produce reliable DAGs. The overall stability of the final $G_{\text{\methodname}}$ also depends on the robustness of the hyperparameter optimization, the ANM-based edge orientation (Section~\ref{ss:directing-edges}), and the cycle resolution mechanism (Section~\ref{ss:final-dag}).

\subsection{Future work} \label{ss:future-work}

As discussed in Section \ref{s:Models-training}, a key direction for future research is the investigation of alternative regressors beyond the deep feed-forward neural networks (DFN) and gradient boosting trees (GBT) used in this study. Both models support SHAP, enabling interpretability of their predictions. However, other machine learning models that also facilitate SHAP explainability could be integrated into the \methodname framework. For instance, Lasso-based methods, such as Group Lasso \cite{yuan2006model}, commonly used in causal discovery approaches like CAM \cite{buhlmann2014cam}, offer inherent sparsity. This property may enhance interpretability by reducing irrelevant connections, potentially yielding more interpretable causal graphs. SHAP compatibility with these methods positions them as promising candidates for future integration into \methodname.

Further exploration could also include models like Elastic Net regression \cite{zou2005regularization}, which combines the sparsity of Lasso with the regularization of Ridge regression, possibly improving feature selection while retaining SHAP explainability. Tree-based ensemble methods such as CatBoost and LightGBM, which have native SHAP support, also present an attractive alternative to GBTs, potentially offering computational advantages on large datasets. Expanding \methodname to incorporate these alternatives could enhance its adaptability and performance across various data structures and application contexts.

Another research area is its application to datasets with distributional shifts, where the data-generating process may vary across different populations or over time. Recent work \cite{Wang2024CausallyInvariantFeatures} has addressed this challenge by identifying features that remain causally invariant across changing distributions. Integrating such principles into the \methodname could enhance its applicability to real-world datasets where the assumption of consistent distributions does not hold, improving its robustness and generalization capability.

Exploring the behavior of the different SHAP values of Equation \eqref{eq:shap-values} in each of the $S$ subsets may provide more information on how the variables are grouped together within a larger causal structure. Relating this future analysis to techniques that give us information about how they are grouped may help us better discern relationships between nodes so that we can distinguish between parents and ancestors.

Additionally, efforts could be made to refine the model’s scalability and enhance its ability to handle larger datasets and higher-dimensional graphs.

While our SHAP discrepancy measure (Section~\ref{ss:final-dag}) offers a novel, explainability-guided heuristic for resolving cycles that may arise from the union of DAGs, a comparative evaluation against other cycle-breaking strategies could be pursued. Future investigations might explore graph-theoretic approaches, such as heuristics for the Feedback Arc Set problem (e.g., \cite{eades1993fast}), or methods adapted from score-based Bayesian network learning to further refine this crucial step.


\section{Conclusion}
    In this paper, we introduced \methodname, a novel causal discovery method that leverages machine learning and Shapley-based explainability to address key challenges in uncovering causal structures, particularly in complex and nonlinear datasets. The approach combines interventional explainers to compute Shapley values with a bootstrapping mechanism, followed by an Additive Noise Model (ANM) for edge orientation. A novel SHAP discrepancy measure further refines the causal graph by quantifying the agreement between Shapley values and their associated features, filtering out spurious relationships.
    
    Our experiments on five families of synthetic datasets, the Sachs single-cell protein-signaling dataset \cite{sachs2005causal}, and the financial decision-making data \cite{causal_debt_datagen} illustrate the effectiveness and robustness of \methodname. Not only does it recover causal relationships with high precision and low orientation error, but it also compares favorably with existing state-of-the-art methods. By providing interpretable insights into the underlying causal dynamics, \methodname enables contrasting inferred causal links with domain expertise.
    
    We highlight three major strengths of \methodname. First, its results are comparable to leading causal discovery methods across varying nonlinearities. Second, the use of Shapley values brings a valuable layer of interpretability, fostering trust and supporting evidence-based decisions. Third, the integration of data-driven modeling, explainability, and a bootstrapping mechanism provides robustness against spurious edges and orientation errors. However, the computational cost of calculating Shapley values in high-dimensional settings remains a key limitation, prompting future work on scalable, approximate solutions and more efficient architectures. Ongoing advances in GPU computing or distributed ML can significantly help to mitigate the scalability limitation in future implementations.
    
    Looking ahead, we see potential for extending \methodname in multiple directions (see Section \ref{ss:future-work}). Investigating its performance under distributional shifts, employing ensemble SHAP strategies to mitigate collinearity issues, and adopting computational frameworks that handle large or streaming datasets will further broaden its applicability. Moreover, integrating domain-specific priors and exploring advanced regressors such as Lasso-based or alternative tree-based ensemble implementations could further enhance performance and interpretability. 
    
    In fields like healthcare, finance, and economics, where reliable causal insights are crucial, \methodname’s interpretability is especially beneficial. We encourage the community to adopt and adapt \methodname for a wide range of applications, and have made all code and datasets publicly available at \url{https://github.com/renero/causalgraph}. Through such collaborations, we anticipate continued progress in interpretable causal discovery, ultimately advancing our ability to unravel complex causal relationships across diverse domains.


\section*{Acknowledgements}
This work has been partially supported by a Ramon y Cajal contract (RYC2019-028578-I), 
a Gipuzkoa Fellows grant (2022-FELL-000003-01), and a grant from the Spanish MCIN (PID2021-126718OA-I00).


\bibliography{manuscript}

\newpage
\begin{appendices}

\section{Experimental validation of Shapley values versus statistical dependence}
\label{app:shapley-validation}

This appendix details the design, implementation, and results of a series of synthetic data experiments specifically conducted to elucidate the relationship between Shapley feature importance scores (as utilized in \methodname) and traditional measures of statistical dependence (marginal and conditional). These experiments complement the broader performance benchmarks of \methodname presented in Section \ref{s:Results} by providing a focused analysis on the behavior of Shapley values in canonical causal structures known to pose challenges for causal discovery algorithms that rely on CI tests. The goal is to empirically validate the theoretical foundations discussed in Section \ref{ss:mathematical-foundation} and address common questions regarding the interpretation of Shapley values in causal contexts, particularly concerning confounders, mediators, colliders, and multicollinearity.

\subsection{Experimental design and methodology}

The experiment was designed to systematically compare Shapley values with statistical dependence measures across various controlled causal scenarios. The key steps are outlined below:

Four fundamental causal structures involving three variables ($X$, $Y$, $Z$, where $Y$ is the target variable for prediction) or two highly correlated parent variables ($X_1, X_2$) were generated (see Table \ref{tab:appA}).

\begin{table}[ht]
\small
\centering
\begin{tabular}{@{}l p{0.75\textwidth}@{}}
\toprule
\textbf{Structure} & \textbf{Description} \\
\midrule
\textbf{Confounder} & $Z \rightarrow X$, $Z \rightarrow Y$. This structure induces a spurious correlation between $X$ and $Y$ due to the common cause $Z$. \\
\textbf{Chain (Mediator)} & $X \rightarrow Z \rightarrow Y$. Here, the influence of $X$ on $Y$ is mediated by $Z$. \\
\textbf{Collider} & $X \rightarrow Z \leftarrow Y$. In this structure, $X$ and $Y$ are marginally independent but become conditionally dependent given their common effect $Z$. \\
\textbf{Collinear Parents} & $X_1 \rightarrow Y$, $X_2 \rightarrow Y$, with $X_1 \approx X_2$. This scenario tests behavior under high multicollinearity between parent nodes. For this structure, $Y$ is the target, and $X_1, X_2$ are the features of interest. \\
\bottomrule
\end{tabular}
\caption{Causal structures considered in the analysis}\label{tab:appA}
\end{table}

For the Confounder, Chain, and Collider structures, the target variable for prediction and Shapley value computation is always $Y$. The features of interest for which Shapley values and dependence measures are reported are $X$ and $Z$.

Data was generated from linear Gaussian structural equations with coefficients set to 1.0, except in the collinear case, where a small perturbation induced multicollinearity. Each dataset comprised $n = 5000$ samples with noise terms drawn from a $\mathcal{N}(0, 0.10^2)$ distribution. The process was repeated $R = 50$ times per structure to ensure robust estimation of means and standard deviations for all metrics.

A gradient-boosted tree model (XGBoost) was fitted to predict the target variable $Y$ from the relevant features ($X, Z$ or $X_1, X_2$, depending on the causal structure). The model hyperparameters were fixed at 200 trees, maximum depth of 4, and learning rate of 0.05 across all experiments to isolate the effects of data structure, despite the linearity of the data generation process.

Shapley values were computed using TreeExplainer from the SHAP library \cite{lundberg2017unified} to quantify feature importance in predicting $Y$. For each of the $R=50$ replicates, mean absolute Shapley values ($\phi$) per feature were calculated on a held-out test set of 1,000 points.

To compare Shapley values with traditional statistical measures, marginal and conditional dependence metrics were computed between each relevant feature and the target $Y$. Marginal dependence was quantified using the Pearson correlation coefficient $\rho(\text{Feature}, Y)$. Conditional dependence was assessed via partial Pearson correlation coefficients, e.g., $\rho(X, Y \mid Z)$ for structures involving additional non-target features. Statistical significance of conditional dependence was evaluated through p-values calculated using Fisher's Z-transform.

\subsection{Aggregated results}
The mean and standard deviation (s.d.) of the computed metrics across the $R=50$ replications are presented in Table \ref{tab:aggregated-results}. Columns are: $\overline{\rho}$ (mean marginal Pearson correlation with $Y$), $\sigma_\rho$ (s.d. of marginal Pearson correlation), $\overline{\rho_p}$ (mean partial correlation with $Y$, conditioned as described above), $\sigma_{\rho_p}$ (s.d. of partial correlation), $\overline{p}$ (mean $p$-value for the partial correlation), $\sigma_p$ (s.d. of $p$-value), $\overline{\phi}$ (mean absolute Shapley value for predicting $Y$), and $\sigma_\phi$ (s.d. of mean absolute Shapley value).

\begin{table}[h!]
\footnotesize
\centering
\caption{Aggregated results (mean $\pm$ s.d. over 50 replications) comparing Shapley values with marginal and conditional dependence measures for different causal structures. The target variable is $Y$. For Confounder, Chain, and Collider, features are $X$ and $Z$. For Collinear, features are $X_1$ and $X_2$.}
\label{tab:aggregated-results}
\resizebox{0.8\textwidth}{!}{
\begin{tabular}{@{}ll r@{}l r@{}l r@{}l r@{}l r@{}l r@{}l r@{}l r@{}l@{}}
\toprule
Structure   & Feature & \multicolumn{2}{c}{$\overline{\rho}$} & \multicolumn{2}{c}{$\sigma_\rho$} & \multicolumn{2}{c}{$\overline{\rho_p}$} & \multicolumn{2}{c}{$\sigma_{\rho_p}$} & \multicolumn{2}{c}{$\overline{p\text{-val}}$} & \multicolumn{2}{c}{$\sigma_{p\text{-val}}$} & \multicolumn{2}{c}{$\overline{|\phi|}$} & \multicolumn{2}{c}{$\sigma_{|\phi|}$} \\
\midrule
Confounder  & X       & 0.990 & & 0.000 & &  0.000 & & 0.013 & & 0.514 & & 0.277 & & 0.356 & & 0.009 & \\
            & Z       & 0.995 & & 0.000 & &  0.706 & & 0.007 & & 0.000 & & 0.000 & & 0.447 & & 0.011 & \\
\addlinespace
Chain       & X       & 0.990 & & 0.000 & & -0.001 & & 0.011 & & 0.625 & & 0.293 & & 0.426 & & 0.009 & \\
            & Z       & 0.995 & & 0.000 & &  0.707 & & 0.006 & & 0.000 & & 0.000 & & 0.372 & & 0.008 & \\
\addlinespace
Collider    & X       & 0.000 & & 0.014 & & -0.990 & & 0.000 & & 0.000 & & 0.000 & & 0.556 & & 0.018 & \\
            & Z       & 0.705 & & 0.008 & &  0.995 & & 0.000 & & 0.000 & & 0.000 & & 0.873 & & 0.023 & \\
\addlinespace
Collinear   & $X_1$      & 0.995 & & 0.000 & &  0.097 & & 0.015 & & 0.000 & & 0.000 & & 0.435 & & 0.012 & \\
            & $X_2$      & 0.995 & & 0.000 & &  0.102 & & 0.015 & & 0.000 & & 0.000 & & 0.361 & & 0.010 & \\
\bottomrule
\end{tabular}}
\end{table}

\subsection{Interpretation and discussion of results}
The results from Table \ref{tab:aggregated-results} provide quantitative evidence supporting the theoretical connection between Shapley values and conditional dependence, as articulated in Section \ref{ss:mathematical-foundation} (particularly Equation \eqref{eq:approximation}).

\subsubsection{Divergence from single-Set CI tests}
The experiments show that Shapley values are not mere proxies for marginal correlation ($\rho$) nor for conditional independence tests based on a single, specific conditioning set (like the partial correlation $\rho_p$ and its $p$-value).

\begin{itemize}
    \item \textbf{Confounder ($Z \rightarrow X,\ Z \rightarrow Y$):}
    For feature $X$, the marginal correlation $\rho(X,Y)$ is very high (mean 0.990) due to the confounder $Z$. However, when conditioned on $Z$, the partial correlation $\rho_p(X,Y|Z)$ is effectively zero (mean 0.000), with a high mean $p$-value (0.514), indicating \( X \ci Y | Z \). Despite this conditional independence given $Z$, the Shapley value for $X$ (mean $|\phi|_X = 0.356$) is substantial. This occurs because, according to Equation (10), Shapley values consider all coalitions $\mathcal{S}$. In coalitions where $Z \notin \mathcal{S}$, $X$ still carries predictive information about $Y$ (as $X$ is a proxy for $Z$'s influence). The non-zero Shapley value reflects this average utility across all contexts. Feature $Z$, the true direct cause of $Y$ (in this simplified model), has a high partial correlation (mean $\rho_p(Z,Y|X) = 0.706$, $p$-value $\approx 0$) and the highest Shapley value (mean $|\phi|_Z = 0.447$).

    \item \textbf{Chain ($X \rightarrow Z \rightarrow Y$):}
    Similarly for feature $X$, $\rho(X,Y)$ is high (mean 0.990). When conditioned on the mediator $Z$, $\rho_p(X,Y|Z)$ is effectively zero (mean -0.001), with a high mean $p$-value (0.625), indicating $X \ci Y | Z$. Yet, $|\phi|_X$ is substantial (mean 0.426). This is because in coalitions $\mathcal{S}$ where $Z \notin \mathcal{S}$, the path $X \rightarrow Z \rightarrow Y$ is "open" in terms of information flow from $X$ to $Y$ for the predictive model, making $X$ valuable. The direct parent $Z$ has a high partial correlation $\rho_p(Z,Y|X)$ (mean 0.707, $p$-value $\approx 0$) and a significant Shapley value (mean $|\phi|_Z = 0.372$).
\end{itemize}
These cases demonstrate that if a standard CI test (like Fisher-Z on partial correlation) indicates conditional independence given a specific variable (e.g., $Z$), the Shapley value for $X$ can still be high. This is because the Shapley value integrates over all possible conditioning sets (coalitions), capturing the feature's predictive contribution in contexts where the specific variable $Z$ is not part of the conditioning set. This aligns with the theoretical premise that Shapley values offer a more global assessment of feature importance.

\subsubsection{Collider structure and conditional dependence}
\begin{itemize}
    \item \textbf{Collider ($X \rightarrow Z \leftarrow Y$):}
    Feature $X$ is marginally independent of $Y$ (mean $\rho(X,Y) = 0.000$). However, when conditioned on the collider $Z$, $X$ and $Y$ become strongly dependent (mean $\rho_p(X,Y|Z) = -0.990$, $p$-value $\approx 0$). The Shapley value for $X$ (mean $|\phi|_X = 0.556$) is high, reflecting this strong conditional dependence that emerges when $Z$ is in the coalition. Feature $Z$ itself, being a child of both $X$ and $Y$ (and thus highly informative about $Y$ when $X$ is known, and vice-versa for the model), has the highest Shapley value (mean $|\phi|_Z = 0.873$). This scenario illustrates that Shapley values correctly identify the importance of variables whose relevance is only revealed upon conditioning.
\end{itemize}

\subsubsection{Robustness and stability in high correlation scenarios}
The Collinear Parents case addresses concerns about behavior with highly correlated variables.
\begin{itemize}
    \item \textbf{Collinear Parents ($X_1 \approx X_2 \rightarrow Y$):}
    Both $X_1$ and $X_2$ are highly correlated with $Y$ marginally (mean $\rho \approx 0.995$). The partial correlations $\rho_p(X_1,Y|X_2)$ and $\rho_p(X_2,Y|X_1)$ are low (means 0.097 and 0.102 respectively), indicating that once one collinear parent is known, the other offers little \textit{unique} additional linear information. Despite these low partial correlations, their $p$-values are effectively zero due to the large sample size, indicating statistical significance.
    Crucially, the Shapley values for $X_1$ (mean $|\phi|_{X1} = 0.435$) and $X_2$ (mean $|\phi|_{X2} = 0.361$) are substantial and intuitively distribute the importance between the two highly correlated features. The standard deviations for these Shapley values are very small ($0.012$ and $0.010$ respectively), highlighting the stability of Shapley attributions in this scenario. This contrasts with the potential instability or difficulty in interpreting individual coefficients in a linear regression model facing severe multicollinearity. This empirical finding supports the discussion in Section \ref{ss:Multicollinearity} regarding the robustness of interventional SHAP explainers.
\end{itemize}

\subsubsection{Confirmation of theoretical connection (Equation (10))}
The results presented along this appendix confirm the theoretical connection established in \ref{ss:mathematical-foundation}, which can equivalently represented as:

\begin{equation}
\phi_j\;=\;\bar\delta_j\times
     \Pr_{\mathcal S\sim w}\!\bigl[X_j\not\!\perp\!\!\!\perp Y\mid\mathcal S\bigr]
\end{equation}

\begin{itemize}
    \item When a feature $X_j$ is strongly and consistently conditionally dependent on $Y$ across many coalitions (high $P_{\text{weighted}}(X_j \nci Y \mid \mathcal{S})$ and/or high average marginal contribution $\overline{\Delta_j}$), its Shapley value is high (e.g., $Z$ in Confounder/Chain, $X$ and $Z$ in Collider, $X_1, X_2$ in Collinear).
    \item When a feature $X_j$'s conditional dependence on $Y$ is only evident in specific contexts (e.g., $X$ in Confounder/Chain is dependent on $Y$ mainly when $Z$ is \textit{not} in the coalition), its Shapley value is moderated but still non-zero, reflecting this averaged importance.
    \item The low standard deviations of the mean Shapley values across 50 runs (typically $<5\%$ of the mean, often much lower) underscore the stability of this importance measure under the experimental conditions.
\end{itemize}

\subsection{Conclusion of experimental validation}
This synthetic study empirically illustrates that Shapley values utilized in \methodname\ capture aspects of feature importance beyond those measured by standard statistical dependence tests. Specifically, Shapley values demonstrate increased robustness under multicollinearity, reveal important distinctions from conventional measures, and substantiate their theoretical relationship to conditional dependence in aggregate. Collectively, these results support the integration of Shapley values into causal discovery frameworks such as \methodname.

\section{Parents selection example} \label{app:feature-selection}

\begin{figure}[ht!]
    \centering
    \begin{minipage}[t]{0.35\columnwidth}
      \subfloat[]{\includegraphics[width=\linewidth]{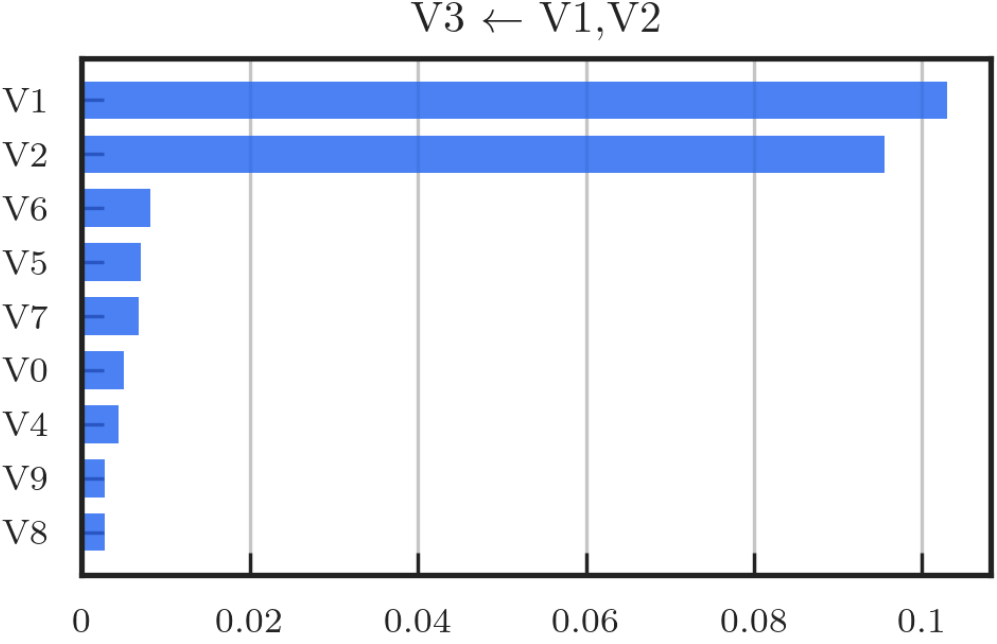}}
    \end{minipage}
    \begin{minipage}[t]{0.35\columnwidth}
      \subfloat[]{\includegraphics[width=\linewidth]{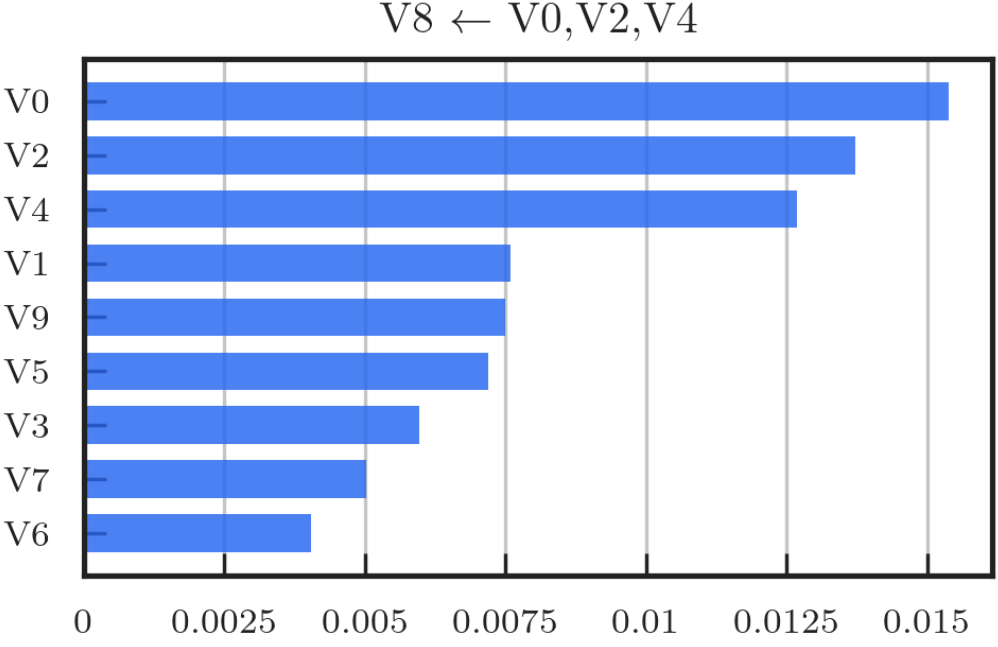}}
    \end{minipage}
    \caption{\footnotesize Example of SHAP values (x-axis) for: (a) endogenous variable $V3$; and (b) $V8$.}
    \label{fig:feature-selection}
\end{figure}

Fig. \ref{fig:feature-selection} illustrates how the parents selection described in Algorithm \ref{alg:parent-selection} mechanism works. On panel (a) there is a significant shift between the features that are more influential ($V1$ and $V2$) and the rest. On panel (${b}$), such shift is not so clear, though the first group of three ($V0, V2$ and $V4$) present a more pronounced influence than the rest. On top of each plot, the variables selected by Algorithm \ref{alg:parent-selection} are written to the right of the target variable, as parents ($\leftarrow$) of the target.

While Algorithm~\ref{alg:parent-selection} is designed for fully automated parent selection, we acknowledge the theoretical possibility of edge cases where its standard DBSCAN-based clustering (detailed in \S~\ref{ss:parents-selection}) might discard features that a user, upon closer inspection of SHAP value distributions (as exemplified in Figure~\ref{fig:feature-selection}), could still deem influential, particularly if SHAP values are very closely clustered or if the automated separation appears too conservative for a specific application. For instance, in a scenario like that depicted in Figure~\ref{fig:feature-selection}b, if the automated process were to hypothetically select only variable $V0$ instead of all three visually distinct top variables ($V0, V2, V4$) due to such an edge case in the clustering, a user might seek further options.

For such specific situations, where an application user might desire finer-grained control or wish to incorporate domain expertise, the concept of ``manual supervision'' is noted here as a practical recourse. This could involve, for example, a domain expert directly reviewing the SHAP value distributions and, based on their judgment, potentially adjusting the automatically selected parent set. As an alternative built-in option for users seeking a more inclusive selection without direct manual intervention, an additional parametrization of Algorithm~\ref{alg:parent-selection} allows it to be run in a \textit{greedy} mode. This mode can be configured to select all features whose SHAP values exceed a user-defined percentile of the distribution, rather than relying solely on the primary DBSCAN clustering outcome. Furthermore, as an optional heuristic, this greedy tuning can be automatically activated if the standard feature selection process results in a causal graph (see \S~\ref{ss:adjacency-matrix-update}) with a number of edges significantly below what might be expected for a densely connected system (e.g., compared to $\frac{p(p-1)}{2}$).

It is crucial to emphasize, however, that the performance and results reported for \texttt{\methodname} throughout this paper (Section~\ref{s:Results}) are based on its fully automated execution, without any such manual supervision or the use of the `greedy mode' or other special parametrizations of Algorithm~\ref{alg:parent-selection}. The discussion of these alternative operational modes is included in this appendix purely as a practical consideration for users who might apply \texttt{\methodname} to new, potentially unique, or particularly challenging datasets where such adjustments could be exceptionally beneficial for fine-tuning the discovery process.

\section{SHAP discrepancy example} \label{app:shap-discrepancy}

\begin{figure}[ht!]
	\centering
	
	\adjustbox{valign=c}{%
		\begin{minipage}{0.33\textwidth}
			\centering
			\subfloat[]{\includegraphics[width=\linewidth]{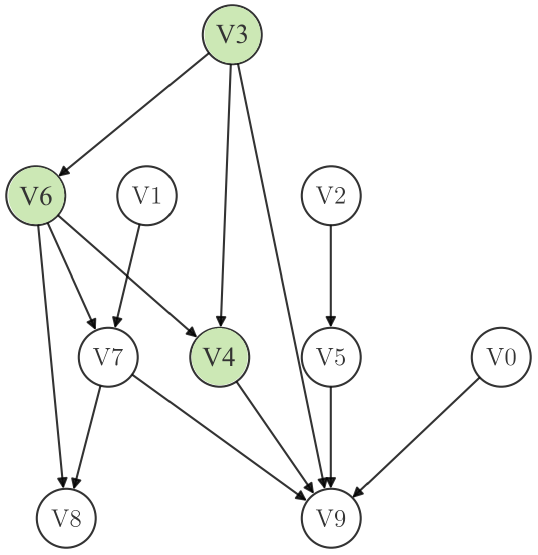}}
		\end{minipage}%
	}\hfill
	\adjustbox{valign=c}{%
		\begin{minipage}{0.65\textwidth}
			\centering
			\subfloat[]{\includegraphics[width=.49\linewidth]{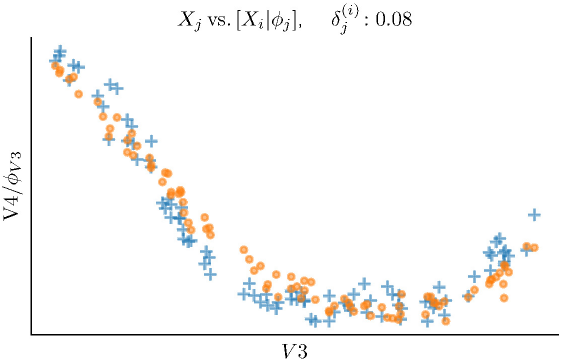}}
			\subfloat[]{\includegraphics[width=.49\linewidth]{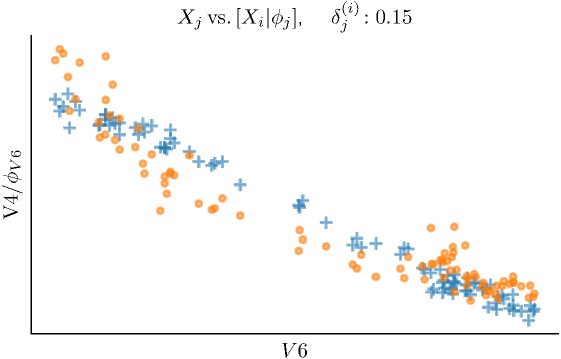}}\\[0.5ex]
			\subfloat[]{\includegraphics[width=.49\linewidth]{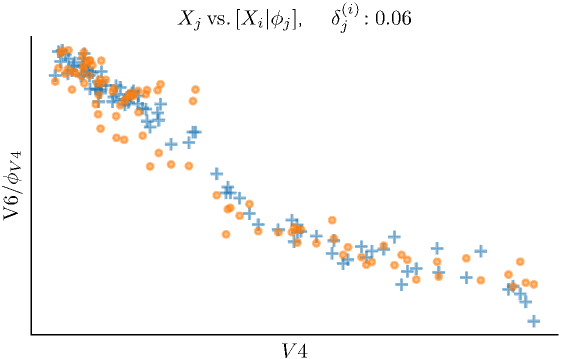}}
			\subfloat[]{\includegraphics[width=.49\linewidth]{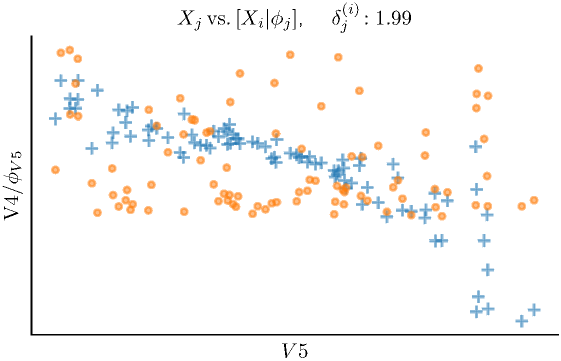}}
		\end{minipage}%
	}
	\caption{\footnotesize ({a}) Ground truth DAG for a system of 10 variables (\(V_0\) to \(V_9\)). Highlighted in green are nodes \(V_3\), \(V_4\), and \(V_6\), which are either direct or indirect causes or effects of one another. (b)-(e) SHAP discrepancy analysis for variable pairs in the graph. As observed, discrepancy values between pairs of \(V_3\), \(V_4\), and \(V_6\) are low ((b), (c) and (d)), in contrast to the one obtained when analyzing the pair \(V_4\), \(V_5\) ({e}).}
	\label{fig:discrepancies}
\end{figure}

Fig. \ref{fig:discrepancies} illustrates the role of SHAP values and the discrepancy between two potential causes or parent features $X_j$, and one effect or dependent feature $X_i$.

In ({b}) we can see discrepancy values for \(V_4\) versus \(V_3\) (\textit{is $V_3$ a direct cause of $V_4$?}), showing a low discrepancy value \(\delta_{ij}^{(t)} = 0.08\), consistent with the fact that \(V_4\) is a direct child of \(V_3\). ({c}) shows that $V_6$ could also be considered a direct cause of $V_4$. However, in ({d}) we can also check that $V_4$ could also be considered a direct cause of $V_6$. In ({e}) we can check how $V_5$ present a very high discrepancy (1.99), indicating that $V_5$ should be discarded among the potential parents of $V_4$.

The low SHAP discrepancy values for causally related variables support the ability of SHAP discrepancy to distinguish causal dependencies, whereas the high discrepancy for non-causal relationships (\(V_4 \rightarrow V_5\)) demonstrates its effectiveness in filtering out irrelevant connections.

\section{Performance comparison between single and combinations of regressors} \label{app:single-vs-union}

\methodname\ employs two complementary regressors—Deep Feed-forward Networks (DFN) and Gradient Boosting Trees (GBT)—to generate individual causal graph hypotheses from the data (see Section~\ref{s:Models-training}). The final DAG, $G_{\text{\methodname}}$, is then constructed by taking the union ($\cup$) of the DAGs derived from each regressor, $G_{DFN}$ and $G_{GBT}$.

The rationale for choosing the union operation is to maximize the discovery of true causal relationships by leveraging the distinct modeling capabilities of DFNs (effective at capturing complex, non-linear functions) and GBTs (robust for tabular data and possessing different inductive biases). Since either type of regressor might uniquely identify certain true edges that the other misses due to its specific learning mechanism, the union aims to create a more comprehensive initial graph, prioritizing the recall of potentially true edges at this stage of the pipeline. It is important to note that each individual graph, $G_{DFN}$ and $G_{GBT}$, is already the result of a robust bootstrapping process (Algorithm~\ref{alg:dag-construction}) designed to stabilize edge selection and reduce spurious connections \textit{before} the union.

We also evaluated the alternative of using the intersection ($\cap$) of $G_{DFN}$ and $G_{GBT}$. The intersection is a more conservative approach, as it retains only those edges agreed upon by both regressors. This can lead to higher precision but may also discard true causal edges identified by only one model type, potentially resulting in a less comprehensive causal graph.

\begin{figure}[t!]
    \centering
    \includegraphics[width=0.4\linewidth]{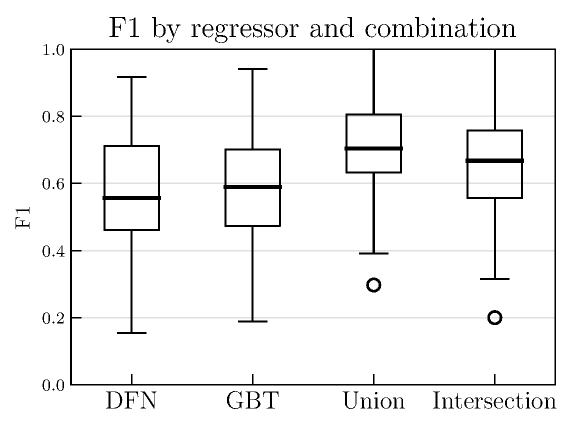}
    \caption{\footnotesize \methodname's F1 scores for all synthetic datasets with 10, 15, 20 and 25 variables, comparing the DAGs obtained by a single regressor (DFN or GBT) to those obtained by combining them through union ($\cup$) and intersection ($\cap$) operations.}
    \label{fig:compare-single-vs-combination}
\end{figure}

Experiments on the synthetic datasets proposed in this study (see Section~\ref{s:Results}) indicate that the union approach results in better overall performance. As shown in Fig.~\ref{fig:compare-single-vs-combination}, the union of the regressors' outputs yields higher F1 scores (0.712) across datasets with 10, 15, 20, and 25 variables compared to both individual regressors (DFN F1: $0.577$; GBT F1: $0.59$) and the intersection strategy (F1: $0.667$). Specifically, the union's F1 score is $23.4\%$ higher than the average F1 score of the individual DFN and GBT models, and $4.9\%$ higher than that of the intersection approach. This suggests that, for the datasets and configurations tested, the union provides a superior balance of precision and recall, capturing more true causal structures overall.

While the union carries a theoretical risk of including more spurious edges than the intersection if one model is prone to specific errors not fully mitigated by its individual bootstrapping phase, the subsequent cycle removal and edge orientation steps (Section~\ref{ss:directing-edges} and Section~\ref{ss:final-dag}), particularly the SHAP discrepancy measure for resolving conflicts, further refine the graph. The empirically superior F1-score ultimately guided our decision to adopt the union approach as the default strategy in \methodname.

\section{Sample DAGs} \label{app:sample-dags}

In this section, a number of sample DAGs are presented to illustrate the differences between the output DAG of the proposed method \methodname and the ones from the comparison methods.

\begin{figure}[t!]
	\centering
	\begin{minipage}[t]{0.575\columnwidth}
		\centering
		\includegraphics[width=\linewidth]{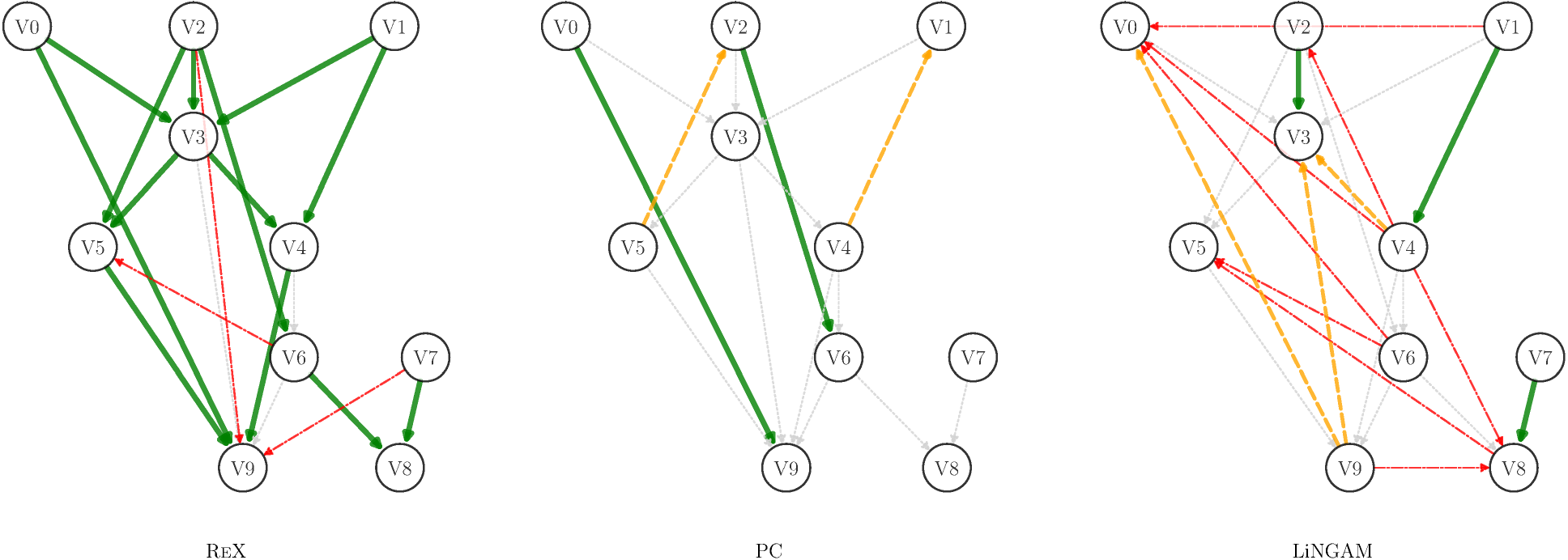}
	\end{minipage}
	\begin{minipage}[t]{0.8\columnwidth}
		\hfill
	\end{minipage}
	\begin{minipage}[t]{0.8\columnwidth}
		\centering
		\includegraphics[width=0.9\linewidth]{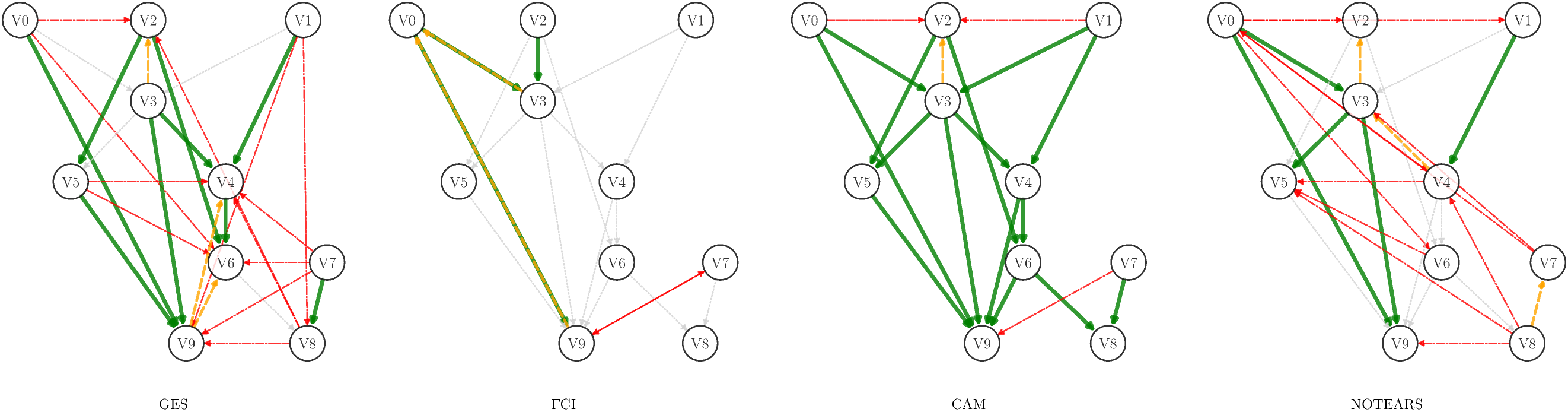}
	\end{minipage}
	\caption{\footnotesize Sample DAGs generated from a Gaussian additive process.}
	\label{fig:gpadd0}
\end{figure}

\begin{figure}[h!]
	\centering
	\begin{minipage}[t]{0.575\columnwidth}
		\centering
		\includegraphics[width=\linewidth]{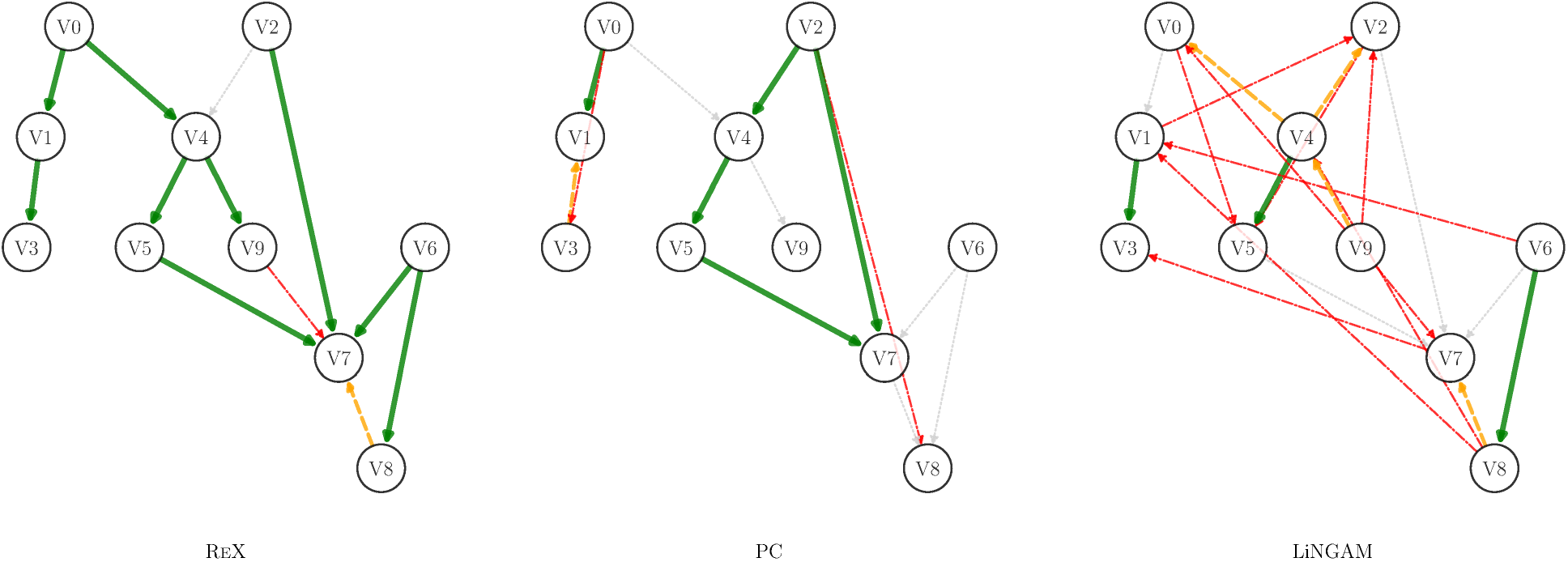}
	\end{minipage}
	\begin{minipage}[t]{0.8\columnwidth}
		\hfill
	\end{minipage}
	\begin{minipage}[t]{0.8\columnwidth}
		\centering
		\includegraphics[width=\linewidth]{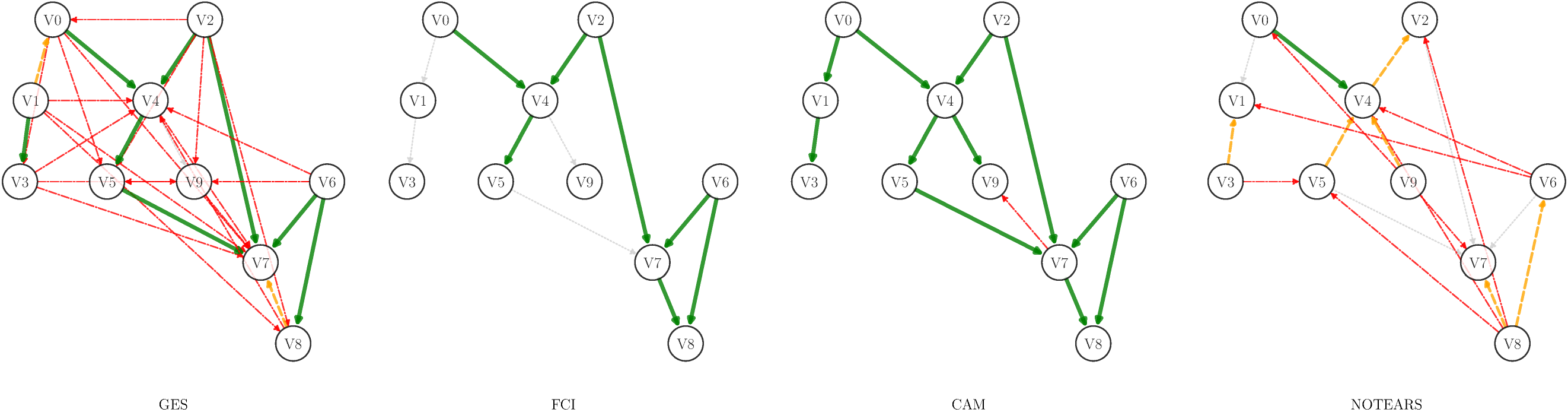}
	\end{minipage}
	\caption{\footnotesize Sample DAGs generated from a Sigmoid additive process.}
	\label{fig:sigadd0} 
\end{figure}

Fig. \ref{fig:gpadd0} and \ref{fig:sigadd0} show sample DAGs generated from a Gaussian additive and Sigmoid additive processes, respectively, with $p=10$ variables. Solid green edges represent correctly predicted causal relationships that match the ground truth, light gray dotted lines indicate edges that were present in the ground truth but not predicted by the method, dashed orange lines represent edges where the causal pair was correctly identified but with the wrong direction, and red dash-dot lines correspond to incorrectly predicted edges.

In Fig. \ref{fig:gpadd0}, \methodname outperforms the other methods by achieving a higher proportion of correctly oriented causal edges (solid green), particularly in areas with more complex relationships, such as nodes \(V_3\), \(V_4\), and \(V_6\). Moreover, \methodname demonstrates a lower number of incorrect predictions (red dash-dot lines) and fewer reversed causal directions (dashed orange) compared to methods like LiNGAM and GES, which show more frequent orientation errors. This highlights \methodname's ability to both recover true causal relationships and accurately predict their direction, even in challenging scenarios.

In Fig. \ref{fig:sigadd0}, \methodname method continues to demonstrate strong performance by recovering a larger number of true causal relationships (solid green lines) compared to other methods. Notably, \methodname accurately captures relationships involving nodes \(V_3\), \(V_4\), and \(V_6\), while methods like PC, LiNGAM, and GES tend to miss several true edges or predict a higher number of reversed causal directions (dashed orange lines). Furthermore, \methodname shows fewer incorrect predictions (red dash-dot lines) relative to CAM and FCI, both of which exhibit a greater number of spurious connections. This highlights \methodname's robustness in discovering complex causal structures and correctly orienting edges, particularly in this dataset where sigmoid additive relationships present a more challenging structure.

\section{Additional results}\label{appendix:add-results}

Here we provide additional results of \methodname for the five considered families of synthetic datasets and the different values of input features $p$.

\begin{figure}[htb!]
	\centering
	\begin{minipage}[t]{0.49\columnwidth}
		\subfloat[]{\includegraphics[width=\linewidth]{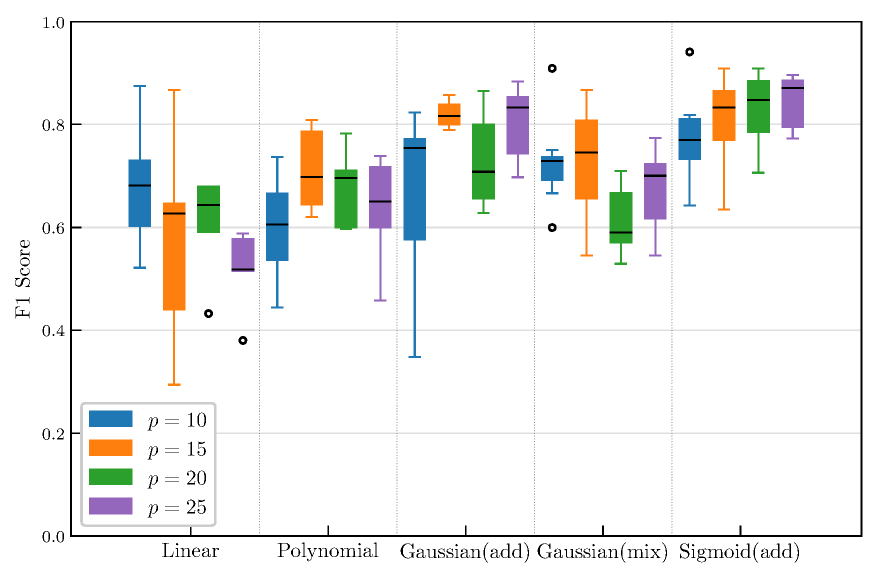}}
	\end{minipage}%
	\begin{minipage}[t]{0.49\columnwidth}
		\centering
		\subfloat[]{\includegraphics[width=\linewidth]{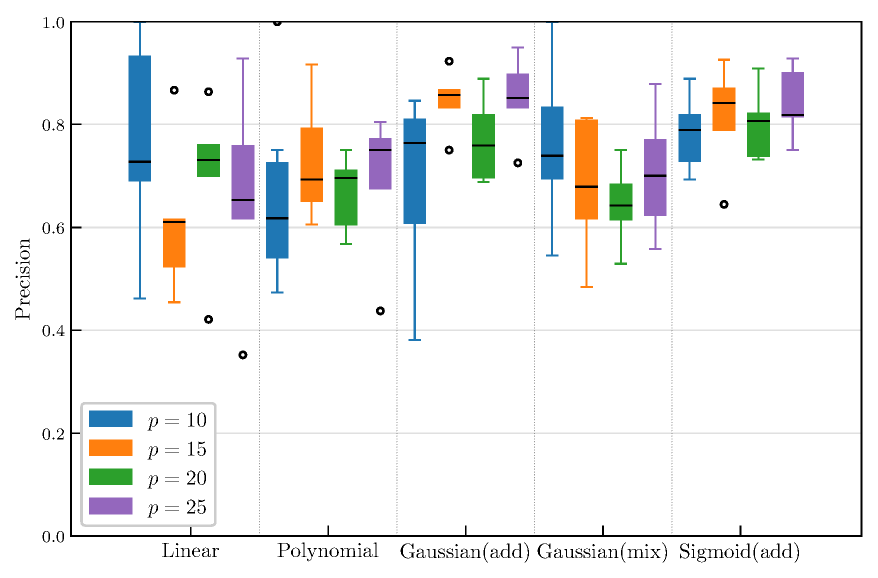}}
	\end{minipage}
	\hfill
	\begin{minipage}[t]{0.49\columnwidth}
		\subfloat[]{\includegraphics[width=\linewidth]{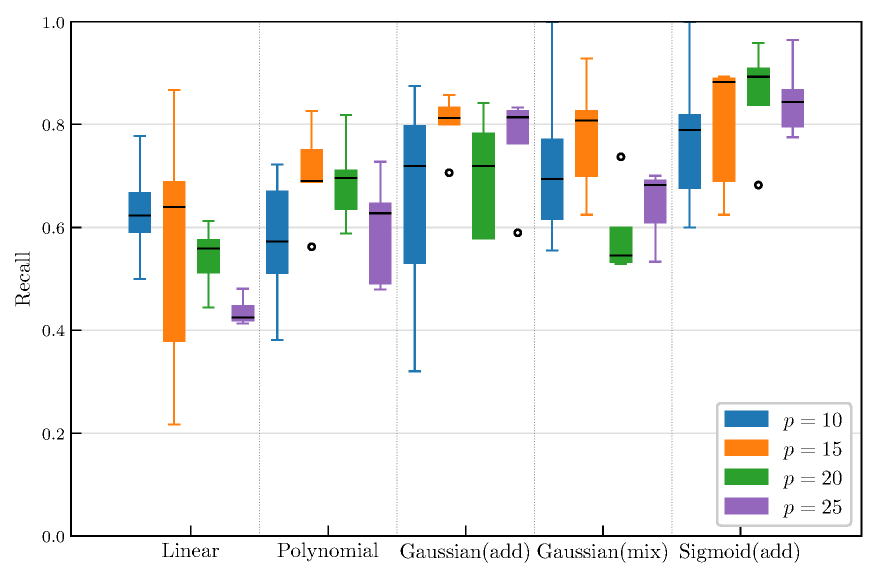}}
	\end{minipage}%
	\begin{minipage}[t]{0.49\columnwidth}
		\centering
		\subfloat[]{\includegraphics[width=\linewidth]{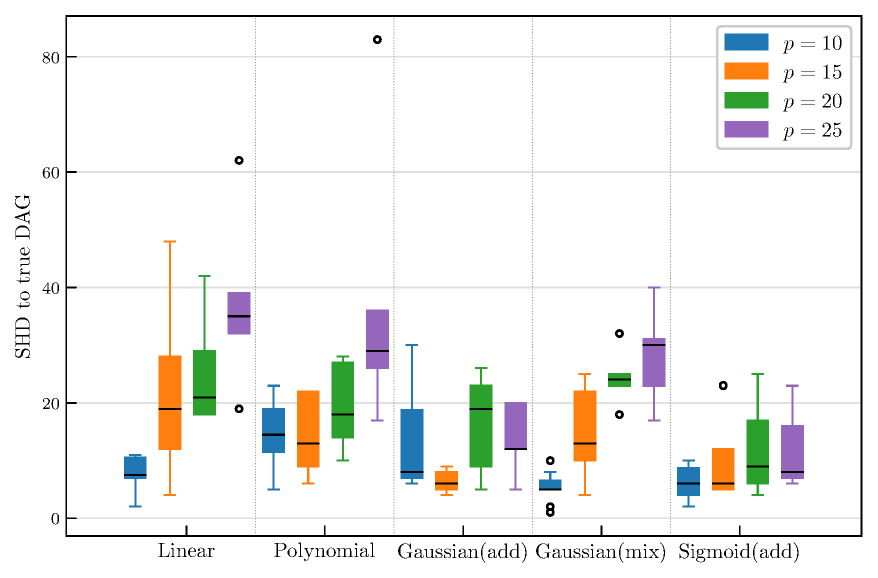}}
	\end{minipage}%
	\hfill
	\begin{minipage}[t]{0.49\columnwidth}
		\subfloat[]{\includegraphics[width=\linewidth]{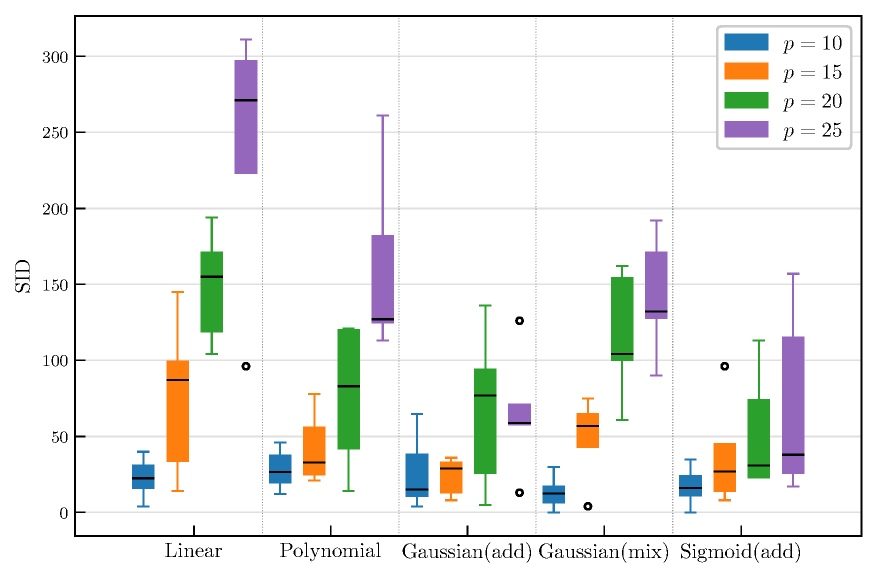}}
	\end{minipage}%
	\begin{minipage}[t]{0.49\columnwidth}
		\centering
		\subfloat[]{\includegraphics[width=\linewidth]{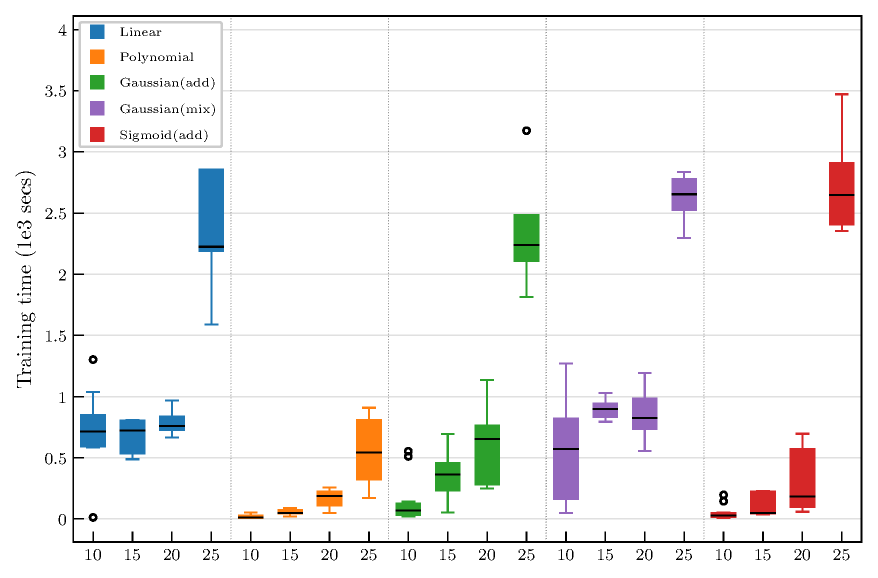}}
	\end{minipage}%
	\caption{\footnotesize 
		F1 (a), Precision (b), Recall (c), SHD (d) and SID (e) for \methodname as a function of the number of features and synthetic data generation mechanism (Appendix~\ref{app:synthetic-datasets}). Panel (f) represents the training time for the different families of synthetic data and number of variables used (x-axis).}
	\label{fig:additional-metrics}
\end{figure}


\section{Synthetic data generation} \label{app:synthetic-datasets}

Each synthetic dataset is built with $p$ variables ($p=10, 15, 20, 25$) and $m=500$ samples.

\begin{enumerate}
  \item The DAG structure is such that the number of parents for each variable is uniformly drawn in $\{0,\ldots,5\}$;
  \item For the $i$-th DAG, the mean $\mu_i$ and variance $\sigma_i$ of the noise variables are drawn as
        $\mu_i \sim \mathcal{U}(-2,2)$ and $\sigma_i \sim \mathcal{U}(0,0.4)$, and the distribution of the
        noise variables is set to $\mathcal{N}(\mu_i,\sigma_i)$;
  \item For each graph, a $500$-sample dataset is i.i.d.\ generated following the topological order of the graph,
        with, for $\ell = 1,\ldots,500$,
        \[
            x^{(\ell)} = \bigl(x_1^{(\ell)},\ldots,x_p^{(\ell)}\bigr), 
            \qquad
            x_i^{(\ell)} \sim f_i\bigl(X_{\operatorname{Pa}(i)},\varepsilon_i\bigr),
            \qquad
            E_i \sim \mathcal{N}(\mu_i,\sigma_i).
        \]
\end{enumerate}

All variables are normalized to zero mean and unit variance.

\medskip
Five categories of causal mechanisms have been considered, replicating those described in \cite{kalainathanGenerativeNeuralNetworks2020}:

\begin{enumerate}[label=\Roman*.]
  \item \textbf{Linear:}
        $X_i = \sum_{j \in \operatorname{Pa}(i)} a_{i,j}\, X_j + \varepsilon_i$, \;
        where $a_{i,j} \sim \mathcal{N}(0,1)$.
        
  \item \textbf{Polynomial:}
        $X_i = \sum_{j \in \operatorname{Pa}(i)} c_{i,j}\, X_j^{d} + \varepsilon_i$, \;
        where $c_{i,j} \sim \mathcal{N}(0,1)$, and $d$ is the degree of the polynomial.
  
  \item \textbf{GP AM:}
        $X_i = \sum_{j \in \operatorname{Pa}(i)} f_{i,j}(X_j) + \varepsilon_i$, \;
        where $f_{i,j}$ is a univariate Gaussian process with a Gaussian kernel of unit bandwidth.
        
  \item \textbf{GP Mix:}
        $X_i = f_i\!\bigl([X_{\operatorname{Pa}(i)},\, \varepsilon_i]\bigr)$, \;
        where $f_i$ is a multivariate Gaussian process with a Gaussian kernel of unit bandwidth.
        
  \item \textbf{Sigmoid Mix:}
        $X_i = f_i\Bigl(\displaystyle \sum_{j \in \operatorname{Pa}(i)} X_j\Bigr) + \varepsilon_i$,
        where $f_i$ is:
        \[
            f_{i,j}(x_j)=
            a \,\frac{b\,(x_j+c)}{1 + b\,(x_j+c)},
            \qquad
            a \sim \operatorname{Exp}(4)+1,\;
            b \sim \mathcal{U}\!\bigl([-2,-0.5]\cup[0.5,2]\bigr),\;
            c \sim \mathcal{U}([-2,2]).
        \]
\end{enumerate}


\section{Computational complexity} \label{ss:computational-complexity}

The computational cost of \methodname, as depicted in Figure~\ref{fig:diagram} (overall workflow) and detailed in Algorithms~\ref{alg:dag-construction} and \ref{alg:parent-selection}, is influenced by several stages. Understanding these helps contextualize the empirical runtime results presented in Figure~\ref{fig:incresing-p}d.

The primary computational stages of \methodname\ are:
\begin{itemize}
    \item \textbf{Model Training (Section~\ref{s:Models-training}):} This involves training $p$ predictive models for each of the two regressor types (DFN and GBT), where $p$ is the number of variables. The cost is $O(p \cdot (C_{\text{HPO}} + C_{\text{train}}))$, where $C_{\text{HPO}}$ is the cost of hyperparameter optimization and $C_{\text{train}}$ is the cost of training a single regressor (DFN or GBT) on $m$ samples and $p-1$ features. This depends on the chosen model's complexity (e.g., network architecture, number of trees) and training epochs.
    \item \textbf{Bootstrapped SHAP Value Computation (Algorithm~\ref{alg:dag-construction}):} This stage iterates $T$ times. In each iteration, for every one of the $p$ target variables, SHAP values are computed for the $p-1$ input features using a subset of $m'$ samples. The cost of computing SHAP values, $C_{\text{SHAP}}$, varies: TreeExplainer for GBTs is generally efficient, scaling with $m'$, $p$, number of trees, and tree depth (e.g., approximately $O(m' \cdot N_{trees} \cdot D \cdot p)$ for $m'$ samples explained over $p$ features). GradientExplainer for DFNs can be more intensive, often scaling with $m'$, $p$, the number of samples for expectation, and the cost of model evaluation. This makes the overall SHAP computation roughly $O(T \cdot p \cdot C_{\text{SHAP}})$.
    \item \textbf{Parent Selection (Algorithm~\ref{alg:parent-selection}):} For each target variable within each bootstrap iteration, this algorithm processes $p-1$ SHAP values. It involves pairwise distance calculations and an iterative DBSCAN clustering approach. Given its current structure involving iterative adjustments of $\zeta$ based on distances, its complexity is estimated to be polynomial in $p$, potentially around $O(p^3)$.
    \item \textbf{Edge Orientation (Section~\ref{ss:directing-edges}):} This is applied to $E_{\text{undir}}$ candidate edges (where $E_{\text{undir}} \leq p(p-1)/2$). It involves fitting two regression models (GAMs are employed for efficiency with $m$ samples) and performing two HSIC tests per edge. HSIC tests typically scale as $O(m^2)$ (or $O(m \log m)$ with approximations). The overall complexity for this stage is roughly $O(E_{\text{undir}} \cdot (C_{\text{GAM}} + C_{\text{HSIC}}))$.
    \item \textbf{Final DAG Construction (Section~\ref{ss:final-dag}):} This includes graph union operations ($O(p^2)$) and cycle detection/resolution. The cost of resolving cycles using SHAP discrepancy depends on the number and complexity of cycles and whether per-sample SHAP values ($\phi_{k,j}$ in Equation~\ref{eq:shap-discrepancy}) are recomputed or readily available.
\end{itemize}

The dominant computational burden in \methodname\ typically arises from the bootstrapped SHAP value computation, especially as $p$ and $m$ increase, leading to a high-degree polynomial scaling in $p$. This theoretical expectation aligns with the empirical runtime trends observed in Figure~\ref{fig:incresing-p}d. Compared to other methods:
\begin{itemize}
    \item \textbf{Constraint-based methods (PC \cite{spirtes2000causation}, FCI \cite{spirtes1999}):} These can be exponential in $p$ in the worst-case due to the need to test conditional independencies over many subsets, though they are often faster for sparse graphs. \methodname\ avoids this exhaustive CI testing by leveraging feature importance from global predictive models.
    \item \textbf{Score-based methods (GES \cite{chickering2002optimal}):} These perform a heuristic search in the space of DAGs, and their complexity depends on the size of this space and the cost of evaluating the score for each candidate structure.
    \item \textbf{SCM-based methods:} LiNGAM \cite{shimizu2006linear} is relatively efficient (e.g., $O(p^3 + p^2m)$) under its specific assumptions (linear, non-Gaussian). CAM \cite{buhlmann2014cam} can be computationally intensive due to its search for parent sets and fitting of non-linear additive models.
    \item \textbf{Continuous optimization methods (NOTEARS \cite{zheng2018dags}):} These typically have polynomial complexity (e.g., $O(mp^2+p^3)$ per iteration or overall for some variants).
\end{itemize}

While \methodname\ incorporates steps that are computationally demanding, particularly SHAP value estimation, its design aims to harness the expressive power of machine learning models and the insights from XAI to uncover complex causal relationships. This represents a trade-off where enhanced accuracy or the ability to handle non-linearities (as demonstrated in Section~\ref{s:Results}) may come at a higher computational cost compared to methods with more restrictive assumptions or less exhaustive feature analysis.

\end{appendices}

\end{document}